%% file: uai2020_main.tex
\documentclass[letterpaper]{article}
\usepackage{uai2020}
\usepackage[margin=1in]{geometry}

\usepackage{times}

\usepackage[utf8]{inputenc} %
\usepackage[T1]{fontenc}    %
\usepackage{url}            %
\usepackage{booktabs}       %
\usepackage{amsfonts}       %
\usepackage{nicefrac}       %
\usepackage{microtype}      %
\usepackage{natbib}

\usepackage{comment}
\usepackage{caption}
\usepackage{subcaption}
\usepackage{paralist}
\usepackage{url}
\usepackage{xspace}
\usepackage{amssymb}
\usepackage{epsfig}
\usepackage{graphicx}
\usepackage{amsmath}
\usepackage{array}

\usepackage{algorithm}
\usepackage{algpseudocode}
\usepackage{xcolor}

\usepackage{multirow}
\usepackage{rotating}
\usepackage{booktabs}

\usepackage{enumitem}
\usepackage{paralist}

\definecolor{citecolor}{HTML}{3498DB}
\definecolor{linkcolor}{HTML}{E74C3C}
\usepackage[pagebackref=false,breaklinks=true,colorlinks,bookmarks=false,citecolor=citecolor,linkcolor=linkcolor]{hyperref}

\usepackage{inconsolata}

\input{space_saver.tex}

\input{definitions.tex}

\newcommand{\supp}[1]{\textbf{\textcolor{red}{SUPP}} \textbf{#1}}

\renewcommand{\supp}[1]{#1}

\title{Analyzing Visual Representations in Embodied Navigation Tasks}

\author{Erik Wijmans\textsuperscript{1, 2, 3}\thanks{~~~Work done while an intern at FRL and FAIR.  Correspondence to \texttt{etw@gatech.edu}} \quad
Julian Straub\textsuperscript{2} \quad
Dhruv Batra\textsuperscript{1, 3} \quad
Irfan Essa\textsuperscript{3} \quad
Judy Hoffman\textsuperscript{1,3} \quad
Ari Morcos\textsuperscript{1} \\
\textsuperscript{1}Facebook AI Research (FAIR) \quad
\textsuperscript{2} Facebook Reality Labs (FRL) \quad
\textsuperscript{3}Georgia Insitute of Technology}

\begin{document}

\maketitle

\input{sections/main/abstract.tex}

\input{sections/main/intro_v2.tex}

\input{sections/main/related_work.tex}

\input{sections/main/approach.tex}

\input{sections/main/initial_exp.tex}

\input{sections/main/init_transfer_exp.tex}

\input{sections/main/rs35_ab_exp.tex}

\input{sections/main/replica_pointnav.tex}

\input{sections/main/discussion.tex}

\input{sections/supplement/supplement_in_main.tex}

\LetLtxMacro{\section}{\oldsection}
\renewcommand{\refname}{\subsubsection*{References}}
{
\bibliography{strings,main}
\bibliographystyle{iclr2020_conference}
}

\clearpage
\appendix
\input{sections/supplement/supplement_content.tex}

\end{document}

%% file: space_saver.tex
\newlength{\sectionReduceTop}
\newlength{\sectionReduceBot}
\newlength{\subsectionReduceTop}
\newlength{\subsectionReduceBot}
\newlength{\abstractReduceTop}
\newlength{\abstractReduceBot}
\newlength{\captionReduceTop}
\newlength{\captionReduceBot}
\newlength{\subsubsectionReduceTop}
\newlength{\subsubsectionReduceBot}

\newlength{\eqnReduceTop}
\newlength{\eqnReduceBot}

\newlength{\horSkip}
\newlength{\verSkip}

\newlength{\figureWidth}
\usepackage{letltxmacro}

\LetLtxMacro{\oldsection}{\section}
\renewcommand{\section}[1]{
    \vspace{-0.05in}
    \oldsection{#1}
    \vspace{-0.05in}
}

\LetLtxMacro{\oldsubsection}{\subsection}
\renewcommand{\subsection}[1]{
    \vspace{-0.02in}
    \oldsubsection{#1}
    \vspace{-0.02in}
}

%% file: definitions.tex
\newcommand{\myquote}[1]{\emph{`#1'}}
\newcommand{\myapprox}{{\raise.17ex\hbox{$\scriptstyle\sim$}}}
\newcommand{\xhdr}[1]{\vspace{0pt}\noindent\textbf{#1}}

\newcommand{\reffig}[1]{Fig.~\ref{#1}}
\newcommand{\refsec}[1]{Sec.~\ref{#1}}

\newcommand{\objectnav}{ObjectNav\xspace}
\newcommand{\turnleft}{\texttt{turn\_left}\xspace}
\newcommand{\turnright}{\texttt{turn\_right}\xspace}
\newcommand{\forward}{\texttt{move\_forward}\xspace}
\newcommand{\callstop}{\texttt{stop}\xspace}

\newcommand{\targetsA}{$\mathcal{A}$\xspace}
\newcommand{\targetsB}{$\mathcal{B}$\xspace}
\newcommand{\aub}{\targetsA $\cup$ \targetsB}
\newcommand{\ie}{\textit{i.e.}\xspace}

%% file: sections/main/abstract.tex
\begin{abstract}
Recent advances in deep reinforcement learning require a large amount of training data and generally result in representations that are often over specialized to the target task. In this work, we present a methodology to study the underlying potential causes for this specialization. We use the recently proposed projection weighted Canonical Correlation Analysis (PWCCA) to measure the similarity of visual representations learned in the same environment by performing different tasks.

We then leverage our proposed methodology to examine the task dependence of visual representations learned  on related but distinct embodied navigation tasks.  Surprisingly, we find that slight differences in task have no measurable effect on the visual representation for both SqueezeNet and ResNet architectures.  We then empirically demonstrate that visual representations learned on one task can be effectively transferred to a different task.
\end{abstract}
\vspace{\abstractReduceBot}

%% file: sections/main/intro_v2.tex
\section{Introduction}

Recent advancements in deep reinforcement learning (deep RL) have allowed for the creation of systems that are able to out-perform human experts on a variety of different games such as Chess, Go, Dota2, and Starcraft2.
These advances have heavily relied on sample-inefficient algorithms that require significant amounts of task-specific training episodes, making them computationally expensive to run.  Furthermore, deep RL has been found to be capable of overfitting to the training task, even for complex problems~\citep{zhang2018study}, or failures when the environment is altered (even if this, in turn, simplifies the task~\citep{ruderman2019safeml}). These observations call into question whether representations learned with one training task will be reusable for novel tasks.

The generality and reuse-ability of representations is a desirable and powerful property as it allows knowledge to be transferred between tasks and can help alleviate a lack of data.
In the regime of supervised learning, it is well known that deep neural networks are capable of overfitting
on tasks and memorizing random labels \citep{zhang2016understanding}, making it reasonable to expect that representations would be highly tuned to their training task.  However, many have shown that representations trained for one task perform well for other tasks, both as an initialization for fine-tuning and as a static feature-extractor \citep{girshick2015fast,anderson2018bottom}. 
Resolving this discrepancy is an area of much debate and active research~\citep{neyshabur2018role,golowich2017size,arora2018stronger,morcos2018importance}.

Reusing representations provides a promising avenue for the emerging field of training virtual robots in simulation before transfer learned skills to reality.
There have been a number of recent works proposing to train robots as Embodied Agents in simulated environments with the ultimate goal of transferring agents learned in simulation to reality \citep{embodiedqa,wijmans2019embodied,habitat19arxiv}.
The ability to reuse representations for new tasks and in new environments is of particular concern to the goal of transferring embodied agents from simulation to reality.  Once in the real world, an agent should be capable of learning new tasks -- such as finding new objects or handling new questions  --  and be able to cope with the non-stationarity of a changing world.  Thus, we seek to answer the following question: \textit{Do different 
embodied navigation tasks induce different visual representations?}

\xhdr{Contributions.}  
First, we adapt the methodologies proposed in \citet{raghu2017svcca} and \citet{morcos2018insights} to examine the impact the task has on the visual representation.  Given two networks, these works  compute their similarity in visual representation by computing the similarity in activations over a set of inputs.

We then study our primary question in the context of the task of Object Navigation (\objectnav), \textit{e.g.} \myquote{Go to the fridge}.  We define two different embodied tasks by constructing disjoint splits of target objects, allowing us to understand the exact differences between our tasks.
We first perform our experiments using SqueezeNet1.2~\citep{iandola2016squeezenet} as parameter efficient networks would be a good choice for embodied agents deployed on real robots.  
We find that, surprisingly, differences in the task do not lead to a measurable effect on the visual representation.
We leverage this knowledge to show that visual representations trained for one tasks are useful for learning another, and, surprisingly, allow for more sample efficient learning.

We then consider how our choice of CNN impacted our findings by performing our analysis on a second CNN, a version of ResNet50~\citep{he2016deep} modified to have a comparable number of parameters to SqueezeNet1.2, and find similar conclusions.

Finally, to evaluate the extent to which these results are environment dependent, we generalize our analysis to multiple permutations of an environment and demonstrate representations learned in one permutation of an environment are effective for the other permutations.

%% file: sections/main/related_work.tex
\section{Related Work}

\xhdr{Representation analysis.}
Analyzing the representations of deep neural networks has been the subject of many works.  Initial works focused on analyzing individual neurons \citep{li2015convergent,arpit2017closer,morcos2018importance}.  In this work, however, we examine the entire representation.  Our closest related works, \citet{raghu2017svcca,morcos2018insights}, propose methods to examine the entire representation of neural networks in the context of standard image classification tasks.  We adopt their analysis tools and utilize them to analyze neural networks in the context of \textit{embodied}-vision tasks and reinforcement learning. See Section \ref{sec:similarity} for a more detailed discussion of the benefits of these methods.

\xhdr{Reward-free reinforcement learning.} 
Transfer of knowledge and representations is a paradigm commonly used in task-agnostic and reward-free reinforcement learning.  The goal of this paradigm is to allow the agent to interact with its environment such that it gains general knowledge, thereby allowing it to learn downstream tasks with less samples.  These works provide the agent with a reward signal such that it will explore its environment (or state-space).  This can be formulated from an information theoretic standpoint to provide intrinsic motivation \citep{jung2011empowerment}.  Others provide a more direct signal in the form of exploration based rewards \citep{burda2018exploration}.  We differ from these works by using representations learned via task-driven reinforcement learning directly for a different task.

\xhdr{Transfer Learning.}  Transfer learning seeks to transfer knowledge between a domain with labeled data to another domain~\citep{pan2009survey}.
Transfer learning has also been studied in the context of reinforcement learning by designing specific objectives or model structures such that knowledge can be transferred between two tasks~\citep{taylor2009transfer}.  We do not use any specific architecture or objectives and examine task dependence of vanilla architectures.

%% file: sections/main/approach.tex
\section{Approach}

In this section, we outline our proposed methodology for examining the effecting of the training task on the visual representation.  We then outline the tasks we will utilize in our experiments.

\subsection{Measuring the similarity of representations} \label{sec:similarity}

In order to examine the impact of the training task on the learned visual representation, we first need a principled way to measure how similarity of two learned visual representations.

A perhaps straight-forward approach to measuring the similarity of representations would be to simply measure the distance (e.g., Euclidean or cosine) between their representations of the same inputs.  However, this approach is ill-suited to neural networks. Consider the following toy example:  For a set of inputs $\mathcal{X}$, suppose that function $f$ produces a representation that is uniform on the N-ball and define $f' = A f$ for an affine transform $A$.  A simple distance calculation (or alternatively, dimensionality reduction and clustering) would report a high distance between the two representations.  Accounting for affine transformations is important when analyzing neural networks as, for any given layer, one can apply any affine transformation to the activations and the inverse to the next layer's weights without changing the network.  Given two neural networks trained in the exact same way modulo the random seed, there is no reason why their representations would be aligned despite computing very similar (if not exact the same) functions~\citep{li2015convergent}.

Instead, we follow the approach of \citet{raghu2017svcca,morcos2018insights} to compare the representations of two deep neural networks.
Given two neural networks, A and B, and a set of $N$ inputs, \citet{raghu2017svcca,morcos2018insights} compare the representations at layer $L$ of both networks by 1) extracting the neuron activation matrix, $X$, of both networks -- where $X_{i, j}$ is the activation of the $i^{th}$ neuron on the $j^{th}$ input; and 2) compute the distance between the neuron activation matrices using Canonical Correlation Analysis (CCA), a classic statistical technique~\citep{hotelling1936relations}.  CCA finds a basis which maximizes the correlation between two matrices and then computes the correlation in that basis, thereby account for any affine transformations between two representations.
It is worth noting that CCA (and variants) do not capture the ``usefulness'' of representation to the downstream task.

We follow the technique proposed by \citet{morcos2018insights} to account for differing numbers of noise dimensions between representations. This method weights CCA correlation coefficients by the amount of variance each CCA direction explains in the real data. Given each of the CCA directions $h_i$ and correlation coefficients $\rho_i$, \citet{morcos2018insights} first computes the projection coefficients 
\begin{equation}
    \alpha_i = \sum_k |\langle d_i, X_k \rangle|
\end{equation}
and then computes 1 minus the weighted average of the correlation coefficients,
\begin{equation}
    \mathbf{D}_{\text{pwcca}} = 1.0 - \frac{1}{\sum_k \alpha_k} \sum_k \alpha_k \rho_k
\end{equation}
as the distance between representations.

\subsection{Measuring the task dependence of learned visual representations.}

Given that we now have a principled way to measure the similarity of learned visual representations, we now outline our proposed method task dependence.

Consider two sets of embodied navigation tasks, \targetsA and \targetsB, that are learnable in the same environment (or set of environments) and contain no overlap, \textit{i.e.} $\mathcal{A} \cap \mathcal{B} = \varnothing$.
A naive approach to using PWCCA to measuring the effect of different task sets on the representation learned would be to train a policy for \targetsA and a policy for \targetsB and then measure the dissimilarity.  Such an approach wouldn't control for the effect of different random initialization, and, more importantly, wouldn't ground the values reported by PWCCA (which is a unitless metric).  Instead, we follow the approach of \citet{morcos2018importance} and compare the distance between models trained on \textit{different} tasks to the distance between models trained on the \textit{same} task. If the distance between models trained on different tasks is \textit{higher} than that between models trained on the same task, representations are \textit{task-dependent} whereas if the distance between models trained on different tasks is the \textit{same} as that between models trained on the same task, representations are \textit{task-agnostic}.

\subsection{Experimental setup}

\xhdr{Task.}
We examine the task of Object Goal Navigation (\objectnav) due to its reliance on both semantic and spatial understanding.  In \objectnav, an agent is given a token describing an object in the environment, such as \textit{fridge}, and then must navigate through the environment until it finds a good view of the fridge and calls the \callstop action.  To avoid under-specification of the task, we restrict target objects to have at most two instances for a given class.  Note that each target object is specified uniquely by its object ID. 

The reward at time $t$ is given as follows:

\[
R_t = \begin{cases}
\frac{\mathbf{\mathbf{IoU}_t}}{\mathbf{IoU}_{\text{max}}} & \text{action = \callstop} \\
-0.05 \cdot  \Delta_{\text{geo\_dist}} & \text{otherwise}
\end{cases}
\]

Where $\mathbf{IoU}$ is the intersection over union between the semantic segmentation of the target object and a predefined bounding box. $\mathbf{IoU}_t$ is the $\mathbf{IoU}$ at the agents current position.  $\mathbf{IoU}_{\text{max}}$ is the maximum possible $\mathbf{IoU}$ for the target object as determined by exhaustive search within a reasonable radius of the target object.

\xhdr{Environment.}
We use the extreme high-fidelity reconstructions in the Replica Dataset~\citep{replica19arxiv} and simlate agents utilizing AI Habitat~\citep{habitat19arxiv}.  We utilize these environment so that our analysis will be more applicable to the ultimate goal of agents operating in reality.  See \reffig{fig:apt} for a top-down view of an environment and the supplement for example agent views. %

\xhdr{Agent.}  The agent has 4 primitive actions,
\forward, which moves $0.25$ meters forward; \turnleft and \turnright (which turn $10$ degrees left and right, respectively),
and \callstop which signals that the agent believes it has completed its task. 
At every time-step, the agent receives an egocentric \texttt{RGB} image and the token specifying the target object.

\xhdr{Policy.}  We parametrize our agent with 3 components.  A visual encoder, a target encoder, and a recurrent policy.  The visual encoder utilizes SqueezeNet1.2~\citep{iandola2016squeezenet} as the backbone architecture as its combination of parameter efficiency and representational power is a logical choice for embodied agents deployed on real robots.
The target encoding is a $128$ dimensional vector that is learn-able and unique for each target object. The policy consists of a GRU~\citep{cho2014learning} followed by 2 fully connected layers. \supp{See the supplementary for more details.}
Note that the vast majority ($\sim$80\%) of the learnable parameters are in the visual encoder.  
The policy and target encoding make approximately $20\%$ of the parameters ($\sim330 \times 10^{3}$ of the total $\sim1.7 \times 10^{6}$).  
This is key to our analysis as otherwise the network is able to perform the task with a frozen randomly initialized visual encoder. 

\xhdr{Training.} 
We use Proximal Policy Optimization (PPO) \cite{schulman2017proximal} 
with Generalized Advantage Estimation \cite{schulman2015gae}. We set the discount factor, $\gamma$, to $0.99$ and $\tau$ to $0.95$.
We collect 128 frames of experience from 32 agents running in parallel (possibly working on different tasks) and then perform 4 epochs of PPO with 2 mini-batches per epoch.  We utilize the Adam optimizer~\cite{kingma2014adam} with a learning rate of $10^{-4}$ and a weight decay of $10^{-5}$.  Note that unlike popular implementations of PPO, we do not normalize advantages as we find this often leads to instabilities during training.  We train for 15,000 rollouts ($\sim 61 \times 10^{6}$ steps of experience) to ensure converge across different random seeds.

%% file: sections/main/initial_exp.tex
\section{How task-depdendent are learned representations?}
\label{sec:pwcca_sim}

\xhdr{Core Hypothesis.} 
Training for different embodied tasks induces different visual representations.  Due to Deep RL's ability to overfit on even complicated tasks, it is reasonable to expect that the representations learned will be highly tuned to their specific task.

\xhdr{Two tasks.} To gain insight into the impact of task differences on visual representations, we must first understand the differences between the tasks themselves. An ideal task set should contain tasks for which the learning and reward dynamics are very similar, but which differ in simple and easily understandable ways. To accomplish this, 
we randomly divide the set of target objects, $\mathcal{X}$, into two equally sized and disjoint subsets \targetsA and \targetsB such that $\mathcal{A} \cap \mathcal{B} = \varnothing$, $\mathcal{A} \cup \mathcal{B} = \mathcal{X}$, and $|\mathcal{A}| = |\mathcal{B}|$ (assuming $|\mathcal{X}|$ is even).  We average our results over three different choices of \targetsA and \targetsB. These two tasks therefore share the same environment and action space and have similar visual statistics, but differ only in the set of target objects to which the agent must navigate.
To control for the effect of any particular environment, we rerun these results over 4 additional environments in the Replica dataset -- \texttt{apartment\_0}, \texttt{office\_2}, \texttt{room\_0}, \texttt{frl\_apartment\_0}.

\begin{figure}[t!]
    \centering
    \includegraphics[width=0.45\textwidth]{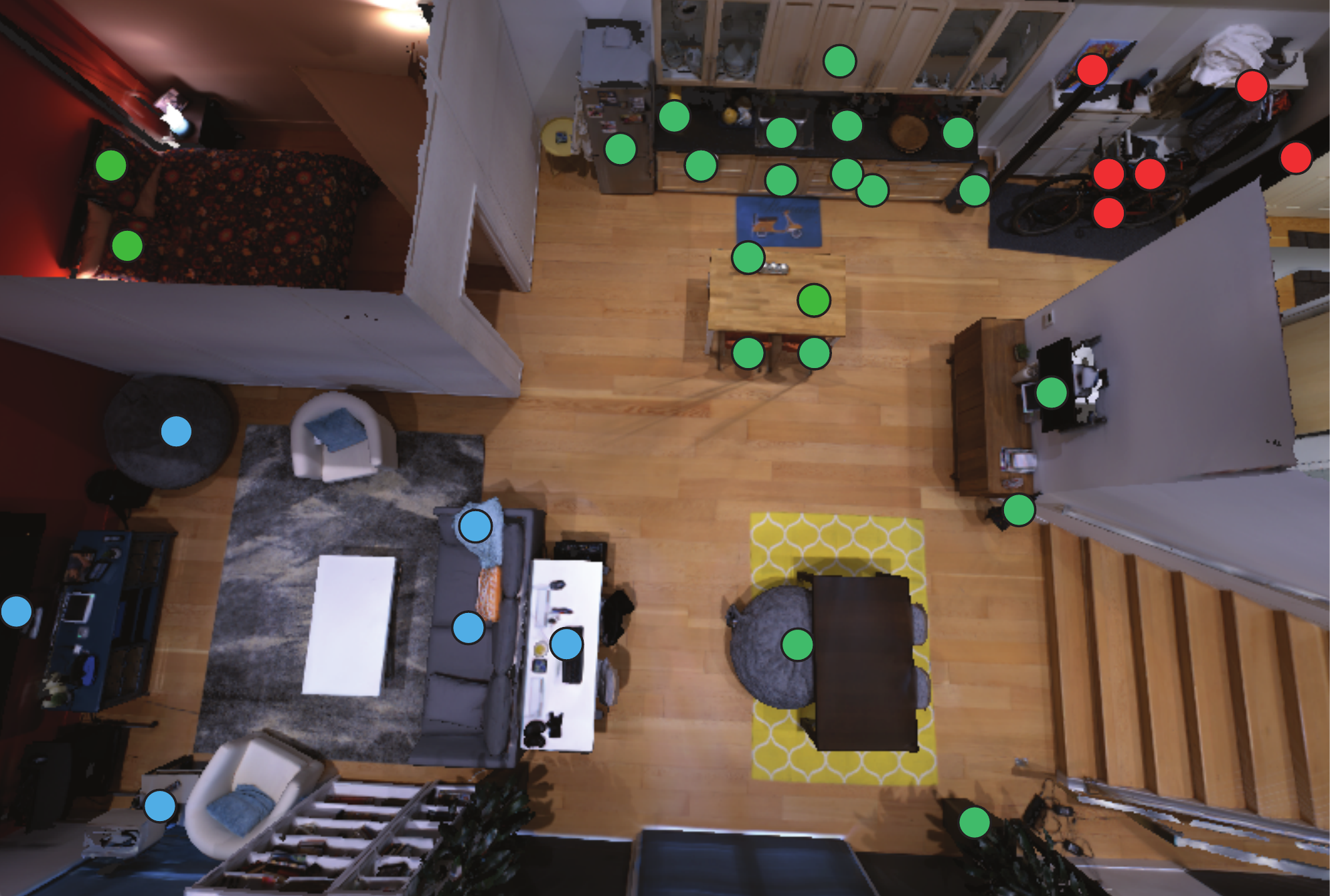}
    \caption{Top-down view in the environment.  Circles denote the location of all target objects.  
}\label{fig:apt}
\end{figure}

\begin{figure}[t!]
    \centering
    \includegraphics[width=\figureWidth]{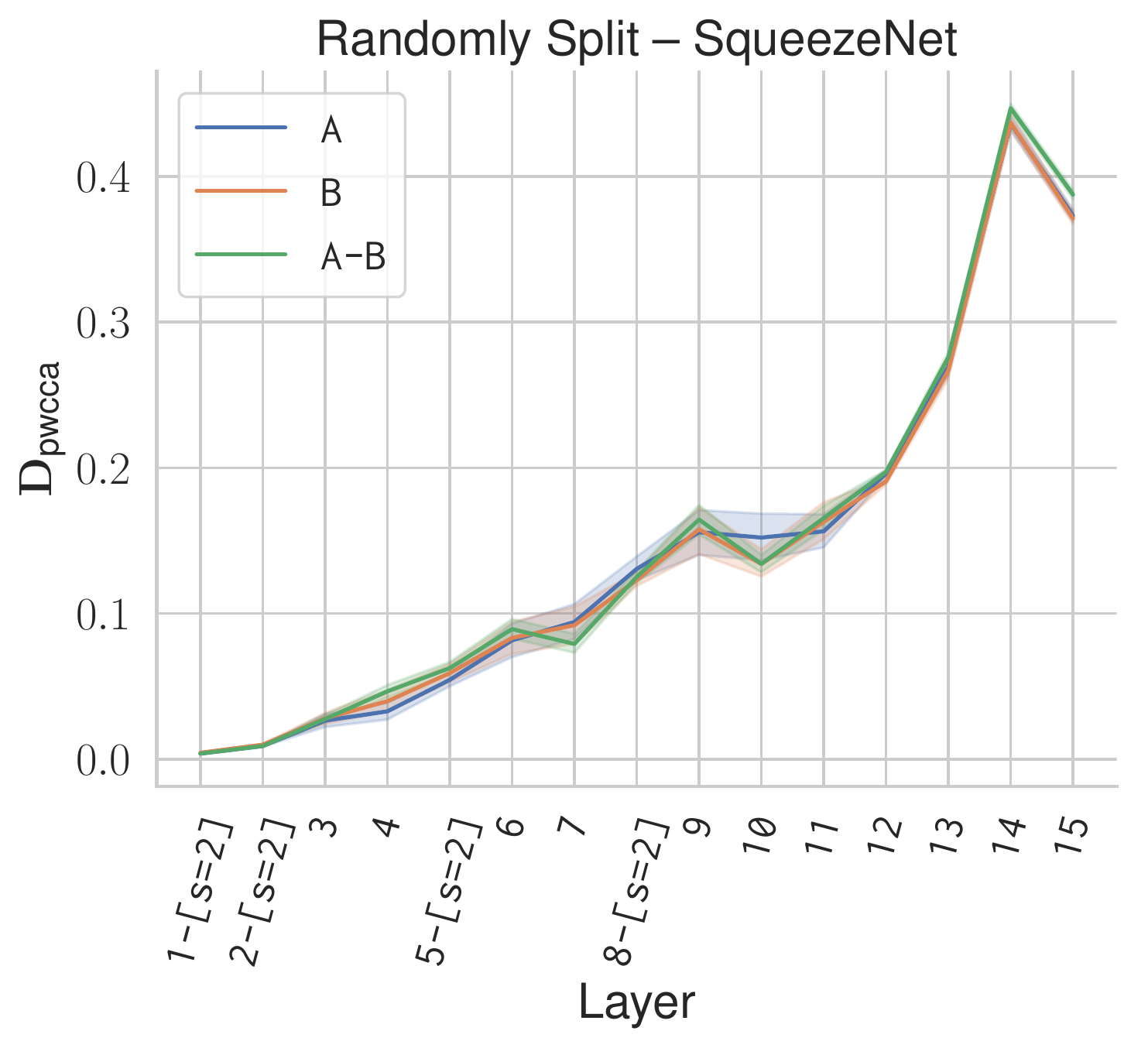}
    \caption{PWCCA results of comparing networks trained on different embodied task.  Surprisingly, the visual representation learned is not influenced by the set of target objects (more than the random seed alone). Down-sampling layers are marked with \texttt{[s=2]}.  Shading around the line corresponds to a $95\%$ confidence interval calculated via empirical bootstrapping.}
    \label{fig:sqz_ab_pwcca}
\end{figure}

To compare representations across different tasks, we train N networks for \targetsA and N networks for \targetsB, compute the PWCCA distance for each pair of networks, and then average over the $N^2$ pairwise comparisons.  To compare representations learned for the same task, we take the N networks trained on \targetsA (or \targetsB), and compute the PWCCA distance for the $\binom{N}{2}$ network pairs.

We use the following notation to denote our comparison: comparisons across networks trained on the same task are denoted without a dash, \textit{e.g.} \texttt{A} is the comparison of networks trained on \targetsA among themselves.  Comparisons across networks trained on different tasks are denoted with a dash, \textit{e.g.} \texttt{A-B} is the comparison \textit{between} networks trained on \targetsA and networks trained on \targetsB.

\begin{figure*}[t!]
\begin{subfigure}{\figureWidth}
    \centering
    \includegraphics[width=\figureWidth]{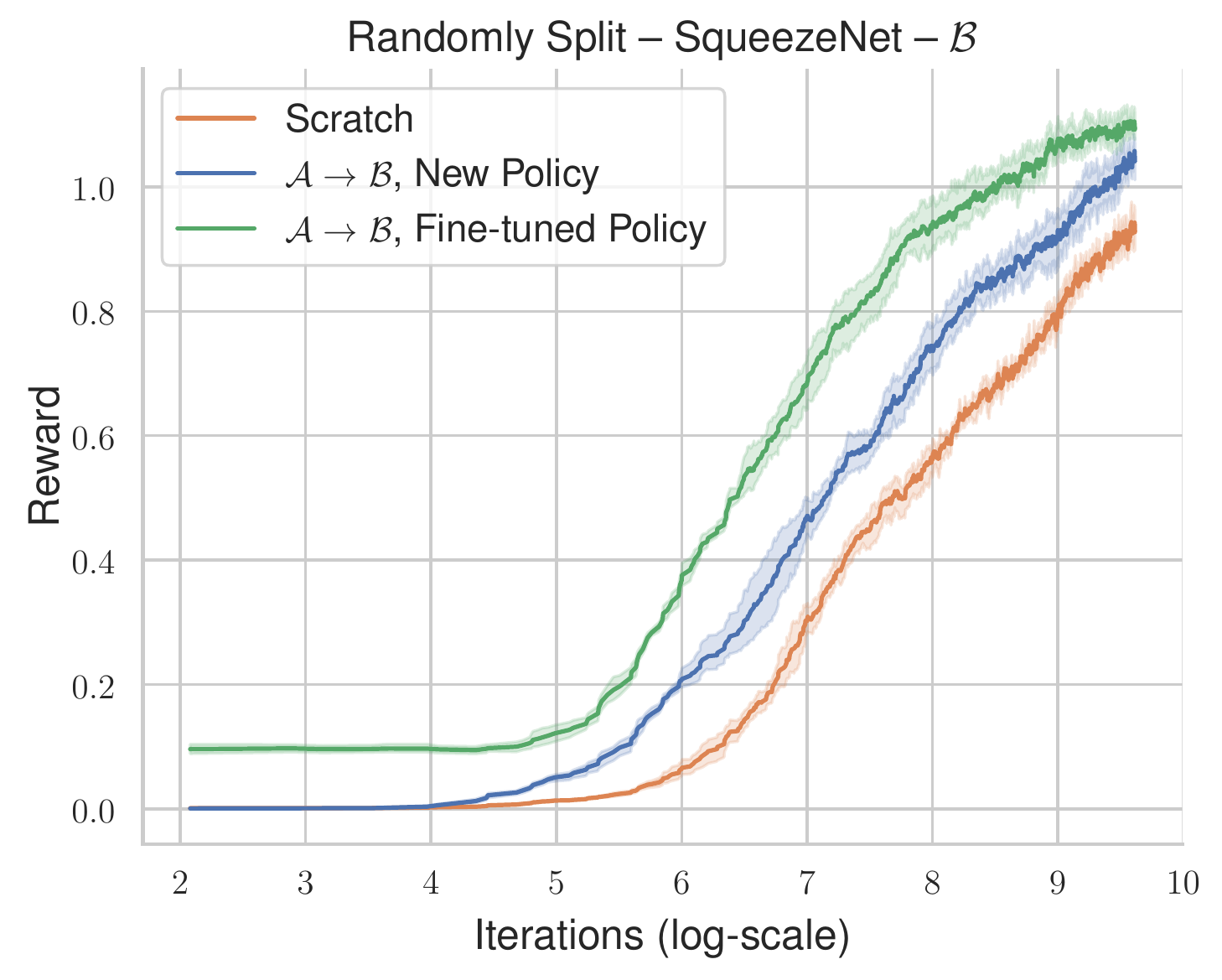}
    \caption{Results of transferring policies learned on \targetsA to \targetsB. We see that the visual representation learned on \targetsA allows for more sample efficient learning of \targetsB.
    }
    \label{fig:transfer_b}
  \end{subfigure}
  \hspace{0.1in}
\begin{subfigure}{\figureWidth}
    \centering
    \includegraphics[width=\figureWidth]{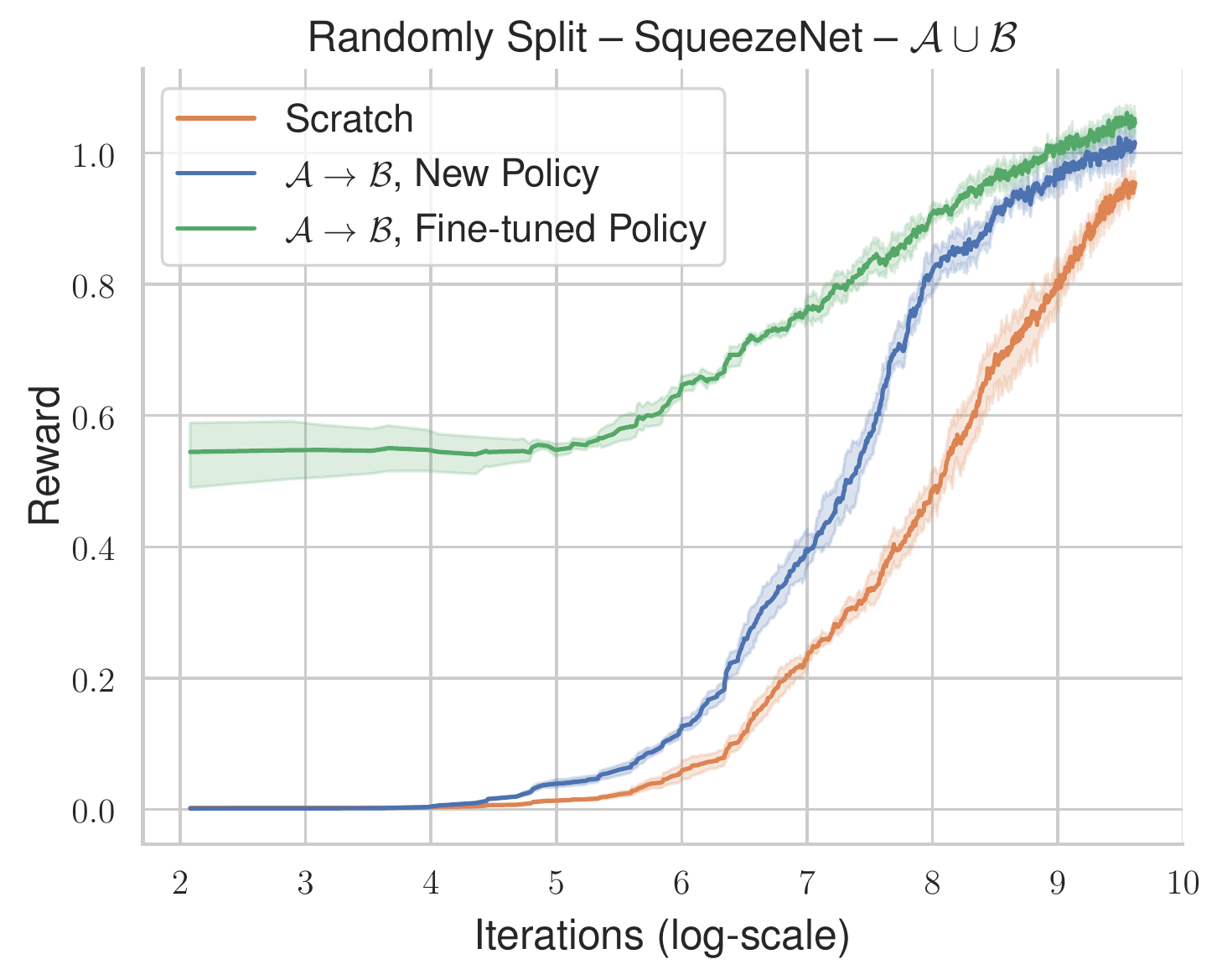}
     \caption{Results of transferring policies learned on \targetsA to \aub.  As in \reffig{fig:transfer_b} we see that the visual encoder learned on \targetsA is well suited to the new task.
    }
    \label{fig:transfer_full}
\end{subfigure}
\end{figure*}

\xhdr{Representations are not influenced by the training task.}
If networks trained on different tasks learn different representations, we would expect the \texttt{A-B} distance to be higher than that for \texttt{A} or \texttt{B} alone. In contrast, we found that distances were similar regardless of task trained, suggesting that networks learn task-agnostic visual representations, \reffig{fig:sqz_ab_pwcca}. 
This result is surprising as it implies that the differences in learning dynamics, reward, and incentives induced by the different target splits \textit{have no more impact on the representation than the random seed alone}.  Despite arising directly from training on that set of target object, the visual representation shows no bias in how it represents the environment. To determine whether this effect is dependent on the particular environment used, we repeated this analysis across four additional environments and found similar trends (\reffig{fig:sqz_random_replica}).
A direct and actionable implication of this result is that the representation learned for one task should transfer to another.

%% file: sections/main/init_transfer_exp.tex
\section{Transferring between \targetsA and \targetsB}

We aim to generate policies with task agnostic visual representations as we hope that these visual representations can be easily adapted to new tasks. In this section, we evaluate whether the PWCCA results above, which suggest that agents learn task-agnostic representations, also imply that representations learned on \targetsA are sufficient to learn \targetsB.

\xhdr{Setup.}
We examine two types of transfer experiments:
1) transferring the policy learned on \targetsA to \targetsB (or from \targetsB to \targetsA), and 
2)  transferring the policy learned on \targetsA to \aub (the full set of targets).
In \textit{all} transfer experiments, every layer of the visual encoder is frozen. We consider both fine-tuning the policy learned on \targetsA and learning a new policy from scratch.

 \begin{figure*}[t!]
     \centering
\begin{subfigure}{\figureWidth}
    \centering
    \includegraphics[width=\figureWidth]{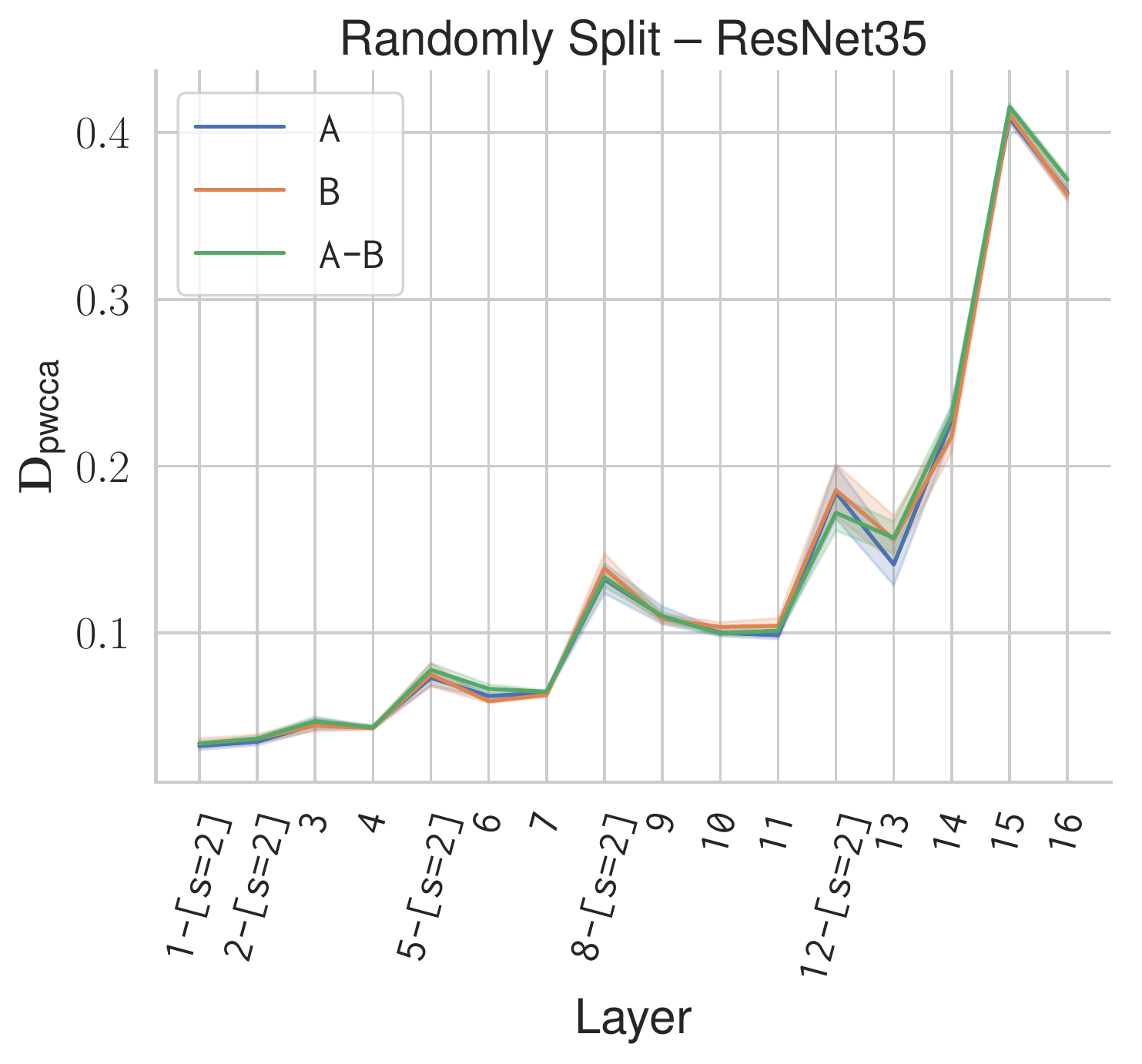}
    \caption{PWCCA results for randomly split target objects for the ResNet35 model.  We find similar trends to SqueezeNet~\reffig{fig:sqz_ab_pwcca}.}
    \label{fig:rs35_ab_pwcca}
\end{subfigure}
\hspace{0.1in}
\begin{subfigure}{\figureWidth}
\centering
    \includegraphics[width=\figureWidth]{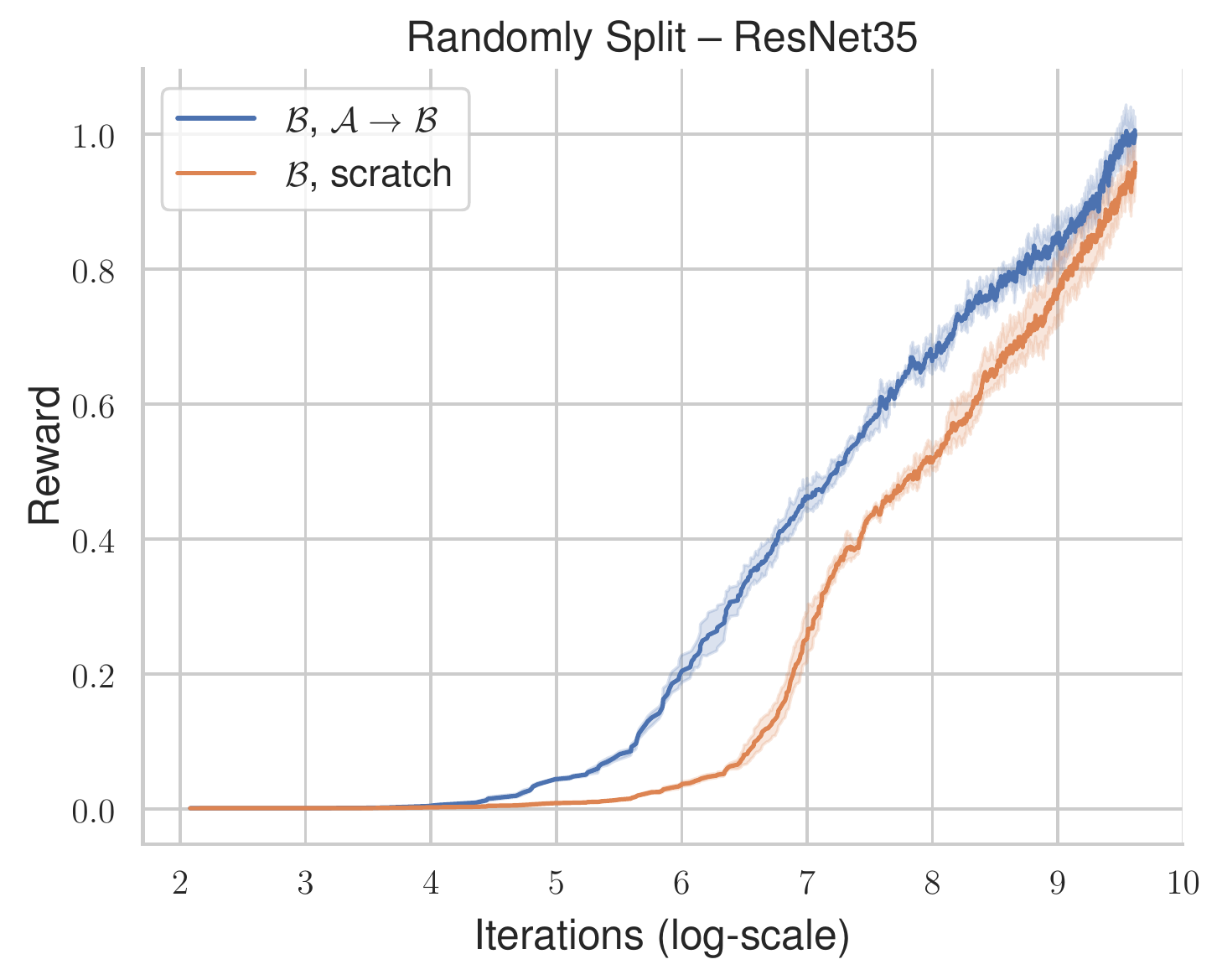}
    \caption{Transfer to \targetsB results for randomly split target objects for the ResNet35 model. Consistent with \reffig{fig:transfer_b}, we see improved sample efficiency when utilizing a \emph{frozen} visual representation learned on \targetsA.}
    \label{fig:rs35_ab_xfer}
\end{subfigure}
 \end{figure*}

\xhdr{Results.}
As suggested by the PWCCA experiments, we found that visual representations learned on \targetsA are effective for learning both \targetsB and \aub (\reffig{fig:transfer_b}, \reffig{fig:transfer_full}). We also found fine-tuning to be more effective than learning a new policy from scratch, suggesting that general navigation skills can transfer in addition to visual representations. 
These results suggest that the representational similarity observed in \refsec{sec:pwcca_sim} leads to directly transferable representations, and confirms that agents in this environment learn task-agnostic representations.

\xhdr{Sample efficient learning of new target objects.}
We also consider the sample efficiency of learning \targetsB with a \textit{frozen} visual representation trained for \targetsA compared to learning \targetsB from scratch. 
Imagine an agent deployed as a home robot: it can be pre-trained for some set of target objects but then must be capable of learning new objects over time.  Ideally we would be able to share and re-use large parts of the agent -- its visual encoder for instance -- to learn these new target objects.   The results in \reffig{fig:transfer_b} imply that using the representation learned on \targetsA as a feature-extractor may provide an efficient method for learning new sets of target objects.

We measure sample efficiency by training with five different random seeds and recording the number of iterations (rollouts) needed to reach a reward of $0.8$ on average, which represents good performance on this task. We compare learning \targetsB from scratch, and learning \targetsB with a \textit{frozen} visual representation pre-trained on \targetsA.  While \reffig{fig:transfer_full} shows that fine-tuning a copy of the existing policy is more sample efficient, we examine the more general case of learning the policy from scratch.

Surprisingly, utilizing a frozen visual representation learned on \targetsA is a more sample efficient strategy for learning \targetsB than learning \targetsB from scratch (\reffig{fig:transfer_b}), suggesting that the visual represnetation learned on \targetsA is generalizable.
We note that learning \targetsB could potentially be done more efficiently as we have not optimized our selection of reinforcement learning algorithm for sample efficiency.  An additional benefit of re-using and freezing the visual encoder is that reinforcement learning algorithms which provide increased sample efficiency but have difficulties scaling to millions of parameters can be used instead.

%% file: sections/main/rs35_ab_exp.tex
\section{A different architecture}

Next, we test how these trends transfer to a different architecture. Specifically, we 
examine a modified ResNet50~\citep{he2016deep} architecture.  We reduce the number of parameters such that the network has a similar number of parameters to SqueezeNet1.2~\citep{iandola2016squeezenet}. The resultant network has 35 layers, and we therefore refer to it as ResNet35. We replace all Batch Normalization layers with Group Normalization~\citep{wu2018group} layers to account for highly correlated observations seen in on-policy reinforcement learning. 
\supp{See the supplementary material for more details}.  We train with the previous procedure and hyper-parameters.

Consistent with SqueezeNet models, we found that randomly distributed target objects have no effect on the visual representation learned (\reffig{fig:rs35_ab_pwcca}, \reffig{fig:rs35_ab_xfer}), indicating that our initial choice of CNN had no impact on this result.  We repeat this analysis over four additional environments and find similar trends (see~\reffig{fig:rs35_random_replica}). We also observe an interesting behavior between the down-sampling layers; the distance between representations induced by different random seeds decreases.  This suggests that residual connections help networks learn more similar representations.
\reffig{fig:rs35_ab_xfer} shows the results of using a representation learned on \targetsA to learn \targetsB.  We once again see that the representation learned on \targetsA is sufficient for learning \targetsB and that this is a more sample efficient strategy than training a network for \targetsB from scratch.

%% file: sections/main/replica_pointnav.tex
\section{Generalization to multiple environments}

\begin{figure*}[t!]
\begin{subfigure}{\figureWidth}
    \centering
    \includegraphics[width=\figureWidth]{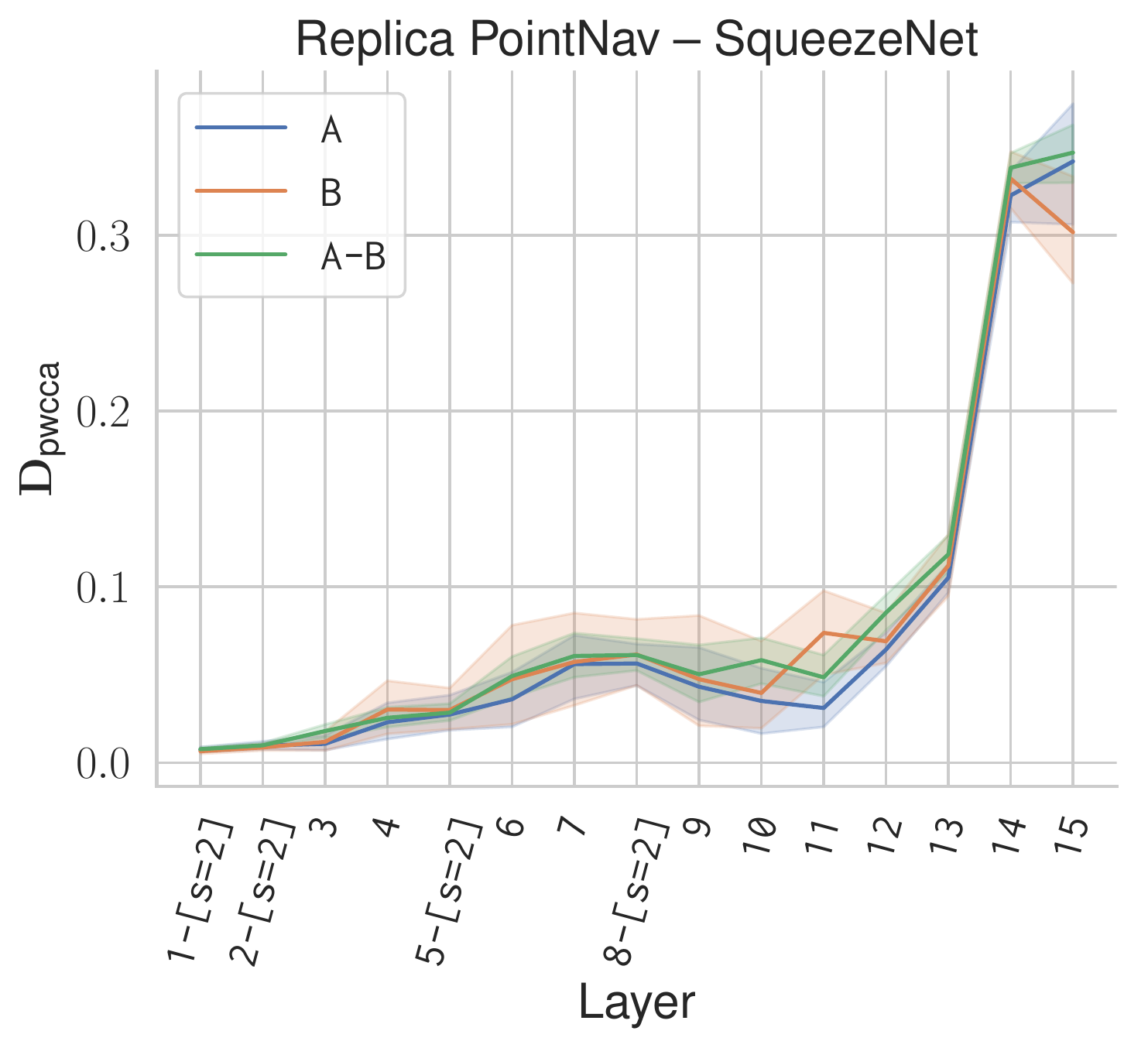}
    \caption{PWCCA results for disjoint sets of training \emph{environments}.}
    \label{fig:pointnav_pwcca_sqz}
\end{subfigure}
\hspace{0.1in}
\begin{subfigure}{\figureWidth}
    \centering
    \includegraphics[width=\figureWidth]{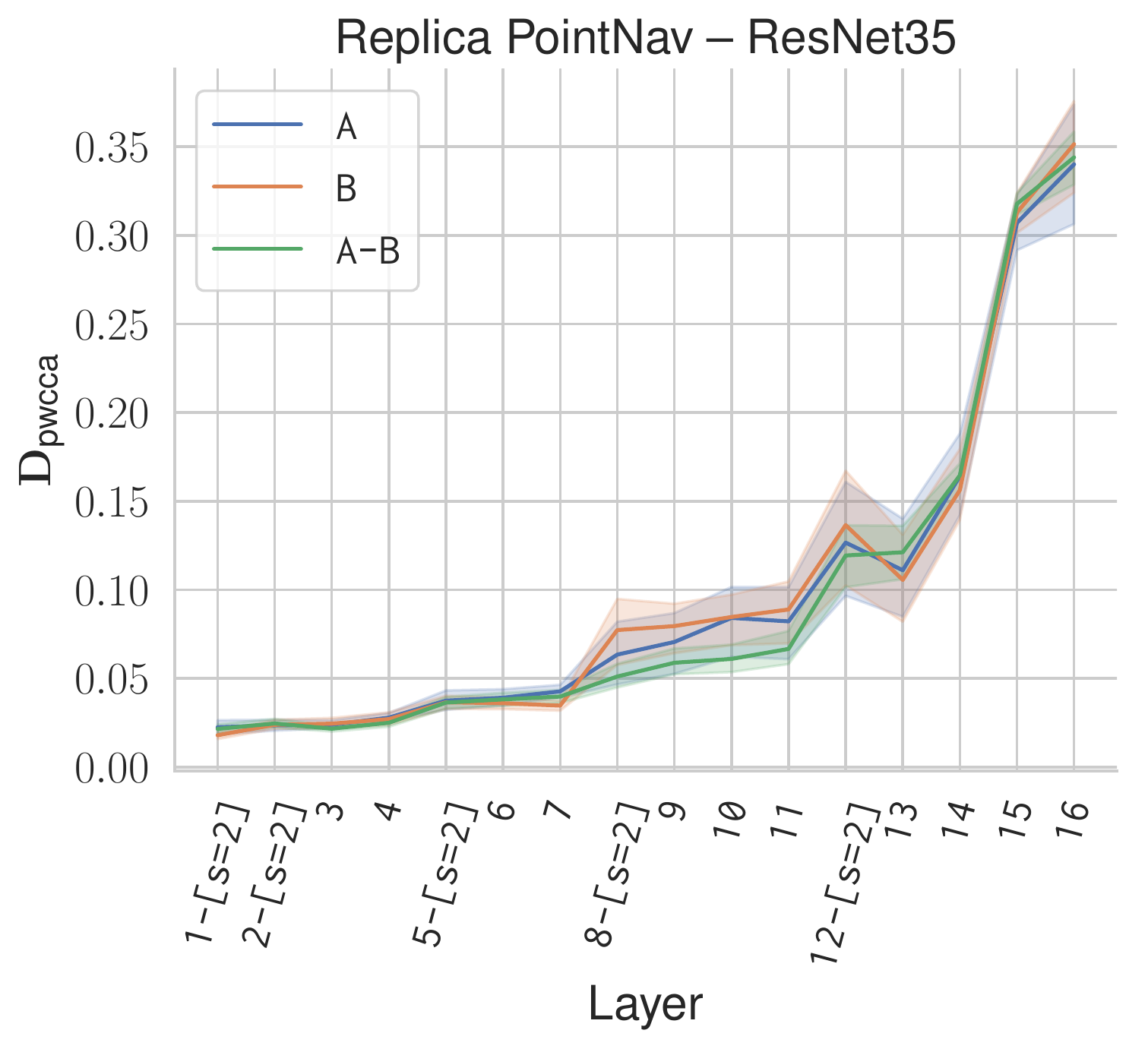}
    \caption{PWCCA results for disjoint sets of training \emph{environments}.}
    \label{fig:pointnav_pwcca_rs35}
\end{subfigure}

\begin{subfigure}{\figureWidth}
    \centering
    \includegraphics[width=\figureWidth]{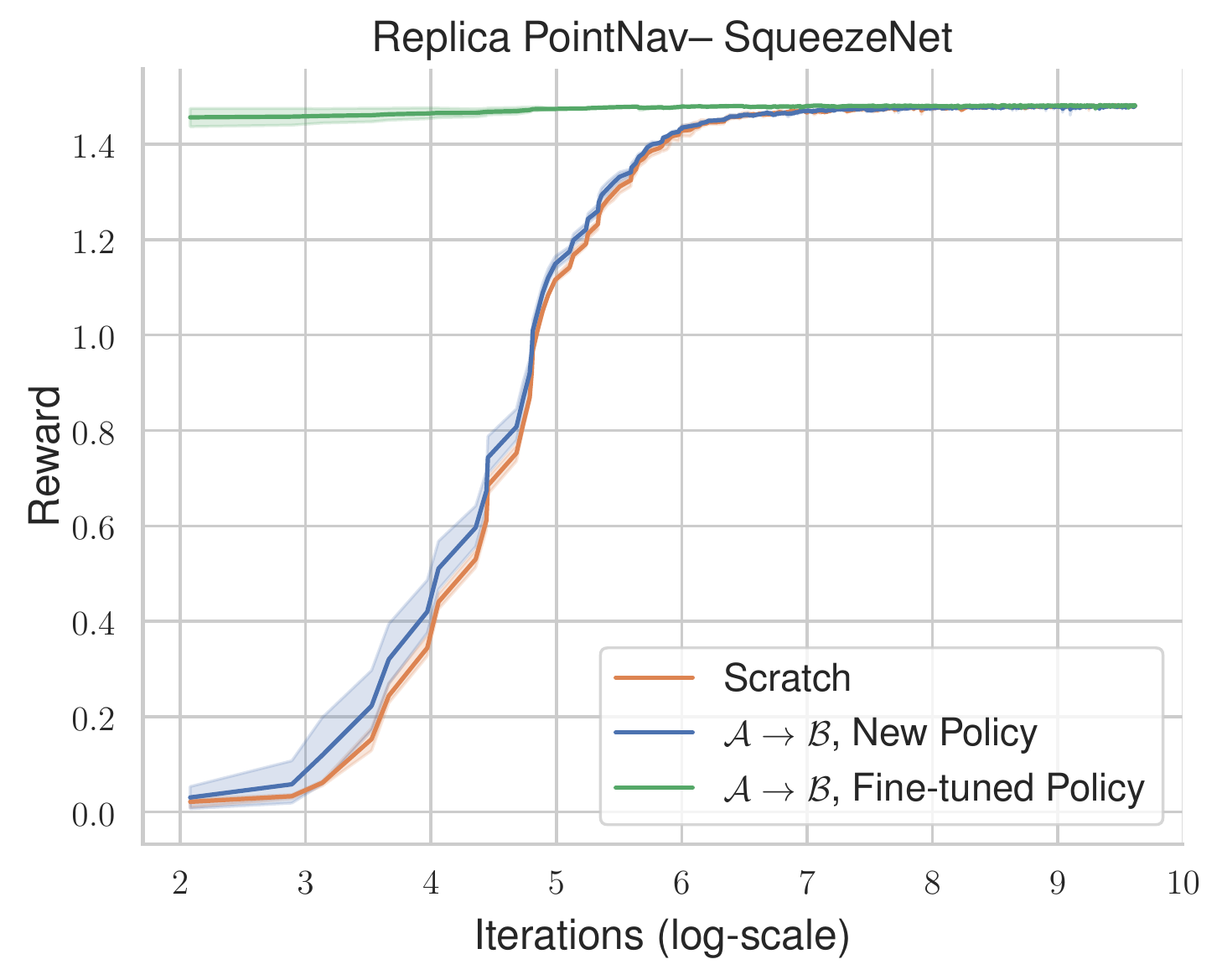}
    \caption{Transfer results for disjoint sets of training \emph{environments}.}
    \label{fig:pointnav_xfer_sqz}
\end{subfigure}
\hspace{0.1in}
\begin{subfigure}{\figureWidth}
    \centering
    \includegraphics[width=\figureWidth]{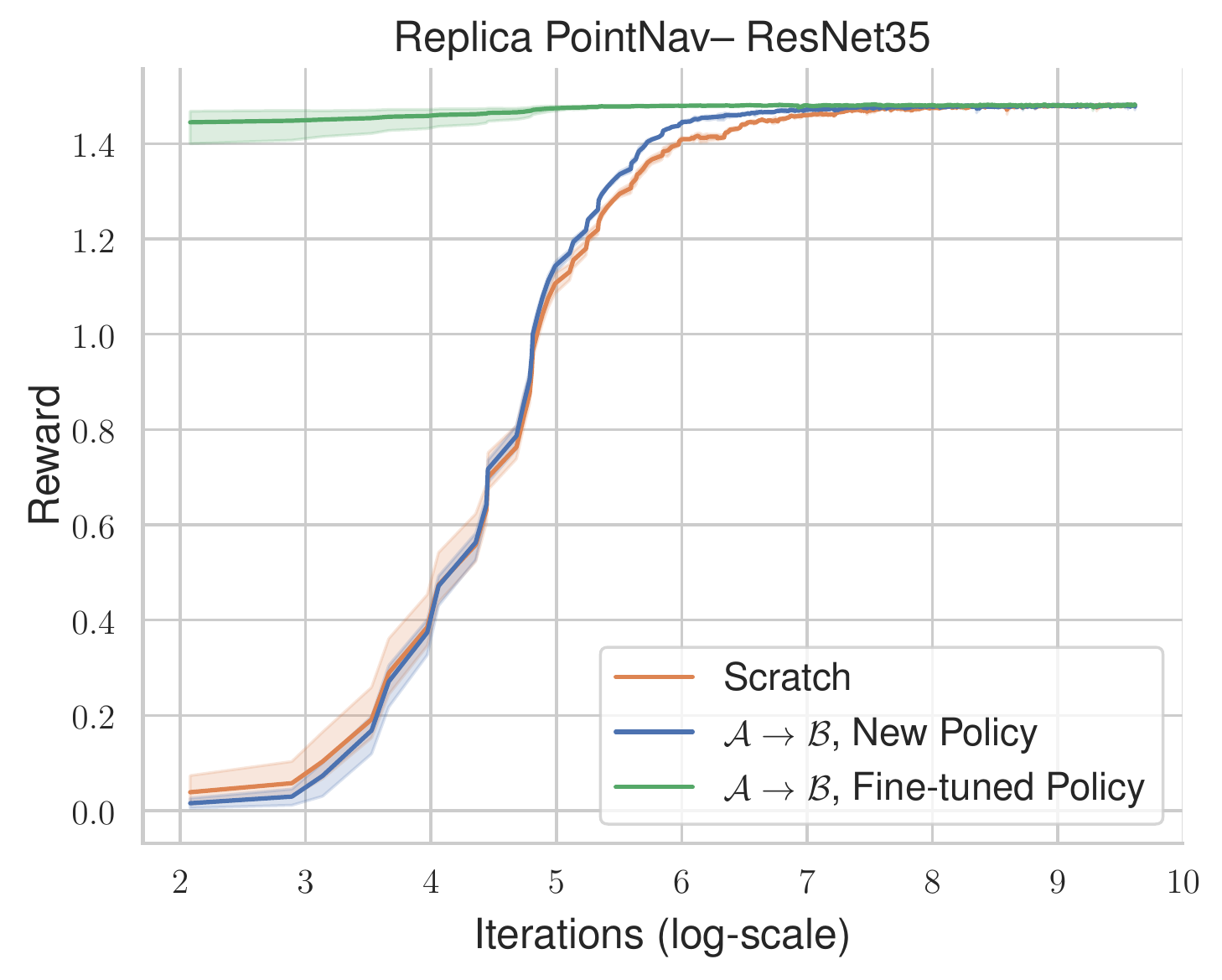}
    \caption{Transfer results for disjoint sets of training \emph{environments}.}
    \label{fig:pointnav_xfer_rs35}
\end{subfigure}
    \caption{Results of training agents in environments with different layouts.  We find that variations in the layout of an environment do not influence the visual representation learned.}
    \label{fig:my_label}
\end{figure*}

Finally, we generalize our analysis to multiple environments.
The Replica dataset contains 6 different version of the same environments with dramatically different configurations of the objects (see \texttt{frl\_apartment\_\{0-5\}}).  This gives us a setting where all environments contain the same objects (\ie table and chairs in various different locations) and structural elements, but have different floor-plans for the agent to navigate.

Utilizing this set of environments we look at the question \textit{Does the representation depend on the position of objects?}.
In order to examine this question, we utilize the task of PointGoal Navigation~\citep{anderson2018evaluation}.  In PointGoal Navigation, the agent must navigation to a point specified in egocentric coordinates (\ie go 5 meters forward and 2 meters to the left).  As in~\citet{habitat19arxiv}, we equip the agent with an RGB camera and GPS+Compass sensor (giving the agent access to its position and orientation relative to its starting location).  We follow the reward structure from~\citet{wijmans2019dd}.
We construct \targetsA as a random selection of 3 environments and \targetsB is the remaining.
We once again find that the representation learned is surprisingly invariant to changes (\reffig{fig:pointnav_pwcca_sqz}, \reffig{fig:pointnav_pwcca_rs35}) -- the location of objects does \emph{not} impact the representation in a measurable way.  We verify this with transfer experiments and find that, in this case, the information learned in environments \targetsA is sufficient to perform the task well in environments \targetsB (\reffig{fig:pointnav_xfer_sqz},\reffig{fig:pointnav_xfer_rs35}).

%% file: sections/main/discussion.tex
\section{Discussion}

We present a series of results and analysis centered around the question:  \textit{Do different embodied navigation tasks induce different visual representations?}  To answer this question, we used PWCCA \citep{raghu2017svcca,morcos2018insights} to measure the influence of the task on the representation, the first to do so for deep RL.
We then constructed two embodied navigation tasks by creating disjoint splits of target objects for the task of \objectnav.
We found that for both SqueezeNet and ResNet visual encoders, the task does not influence visual representation, allowing for use in learning new tasks in a sample efficient manner.

\xhdr{Caveats.} Our results and analysis have the following primary caveat.  The two tasks we examine, while distinct, are quite similar.  Designing experiments for this type of analysis across tasks with less similarity while not introducing too many additional variables is an avenue for future work.

\xhdr{Takeaways.}  We show that under certain settings, task agnostic visual representation can be induced.  Our results suggest that on ingredient is coverage of the visual space that will be seen, implying that designing tasks and environments which maximize the visual diversity seen by the agent is paramount.

%% file: sections/supplement/supplement_in_main.tex
\renewcommand\thesection{\Alph{section}}
\setcounter{section}{0}
\renewcommand\thefigure{A\arabic{figure}}
\renewcommand\thetable{A\arabic{table}}
\setcounter{figure}{0}
\setcounter{table}{0}
\phantomsection

\section{Architecture Details}

\subsection{SqueezeNet Encoder}

For our SqueezeNet1.2~\citep{iandola2016squeezenet} based visual encoder, we utilize all layers expect for the final convolution and global average pool.  Given a 224$\times$224 image, this produces a (512$\times$13$\times$13) feature map.  We follow this with two convolution layers, \texttt{Conv-[42, k=3, d=2]}, \texttt{Conv-[21, k=3, d=1]} where \texttt{d} specifies the dilation, to produce a (21$\times$7$\times$7) feature map.  This feature map is then flattened and transformed to a 256d vector with a fully connected layer.

\subsection{ResNet Encoder}

We construct ResNet35 from ResNet50~\citep{he2016deep} as follows -- We start with all residual layers (all layers minus the global average pool and sofmax classifier).  We then reduce the number of output channels at each layer by a factor of 4.  We then remove 1 residual block within each layer and remove an additional residual block in the 3rd layer.  ResNet50 contains 3 blocks in the first layer, 4 in the second, 6 in the third, and 3 in fourth.  Our ResNet35 contains 2 blocks in the first layer, 3 in the second, 4 in the third, and 2 in the fourth.  We replace all Batch Normalization layers with Group Normalization~\citep{wu2018group} layers, to account for the highly correlated observations seen in on-policy reinforcement learning.

We reduce the 512$\times$7$\times$7 feature map to 41$\times$5$\times$5 with two convolution layers, \texttt{Conv-[41, k=1]}, \texttt{Conv-[41, k=3]}.  This feature map is then flattened and transformed to a 256d vector with a fully connected layer.

\subsection{Policy}

Given the 256-d visual feature, we concatenate the 128-d target encoding and use the resulting 384-d vector as input to a single layer GRU~\citep{cho2014learning} with a $256$-d hidden state.  The hidden state is reduced to 128-d with a fully connected layer, and the 128-d representation is used to produced the softmax distribution over the action space and estimate the value function.

\section{Implementation Details}

Models are trained on a single node with 8 Tesla V100 GPUs.  We use PyTorch to train our agent.

We utilize the publicly available implementation of PWCAA~\citep{morcos2018insights}: \url{https://github.com/google/svcca}.

%% file: sections/supplement/supplement_content.tex
\clearpage

\begin{figure}
    \centering
    \includegraphics[width=0.45\textwidth]{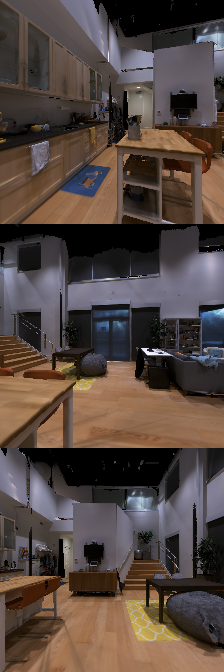}
    \caption{Example images from the environments we utilize.}
\end{figure}

\begin{figure}
    \centering
    \includegraphics[width=0.45\textwidth]{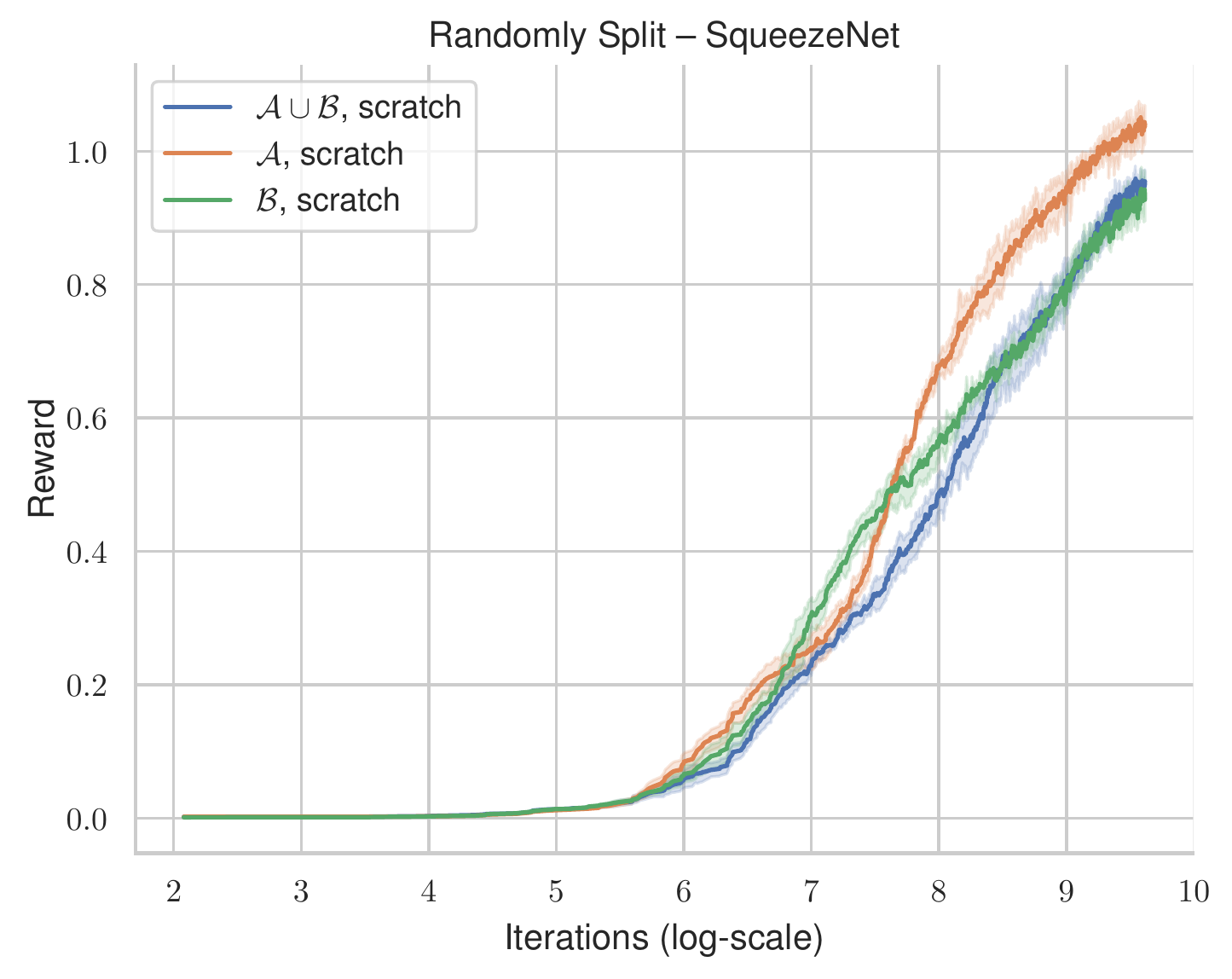}
    \includegraphics[width=0.45\textwidth]{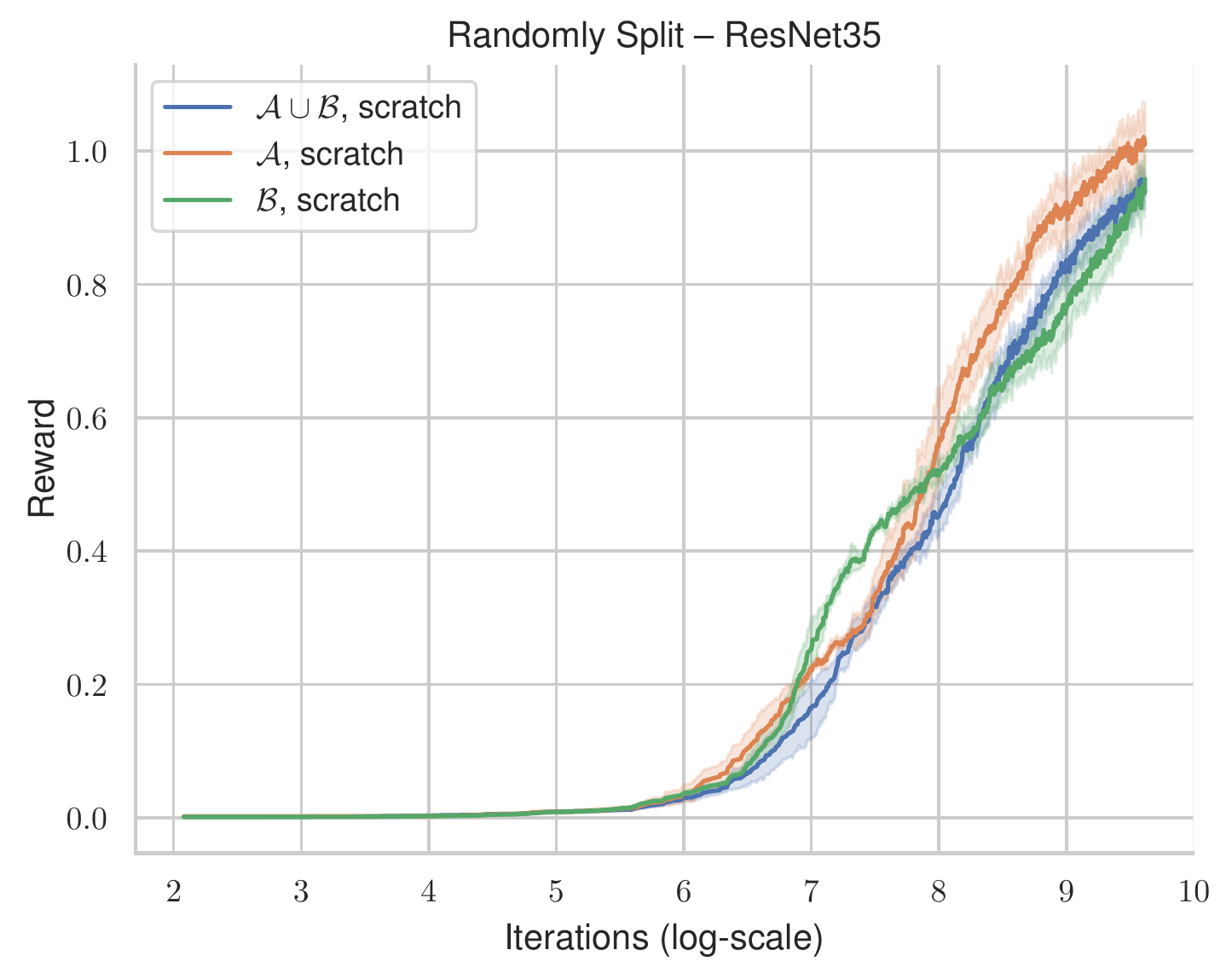}
    \caption{Reward curves for both on \targetsA. \targetsB, and \aub.}
\end{figure}

\begin{figure}
    \centering
    \includegraphics[width=0.5\textwidth]{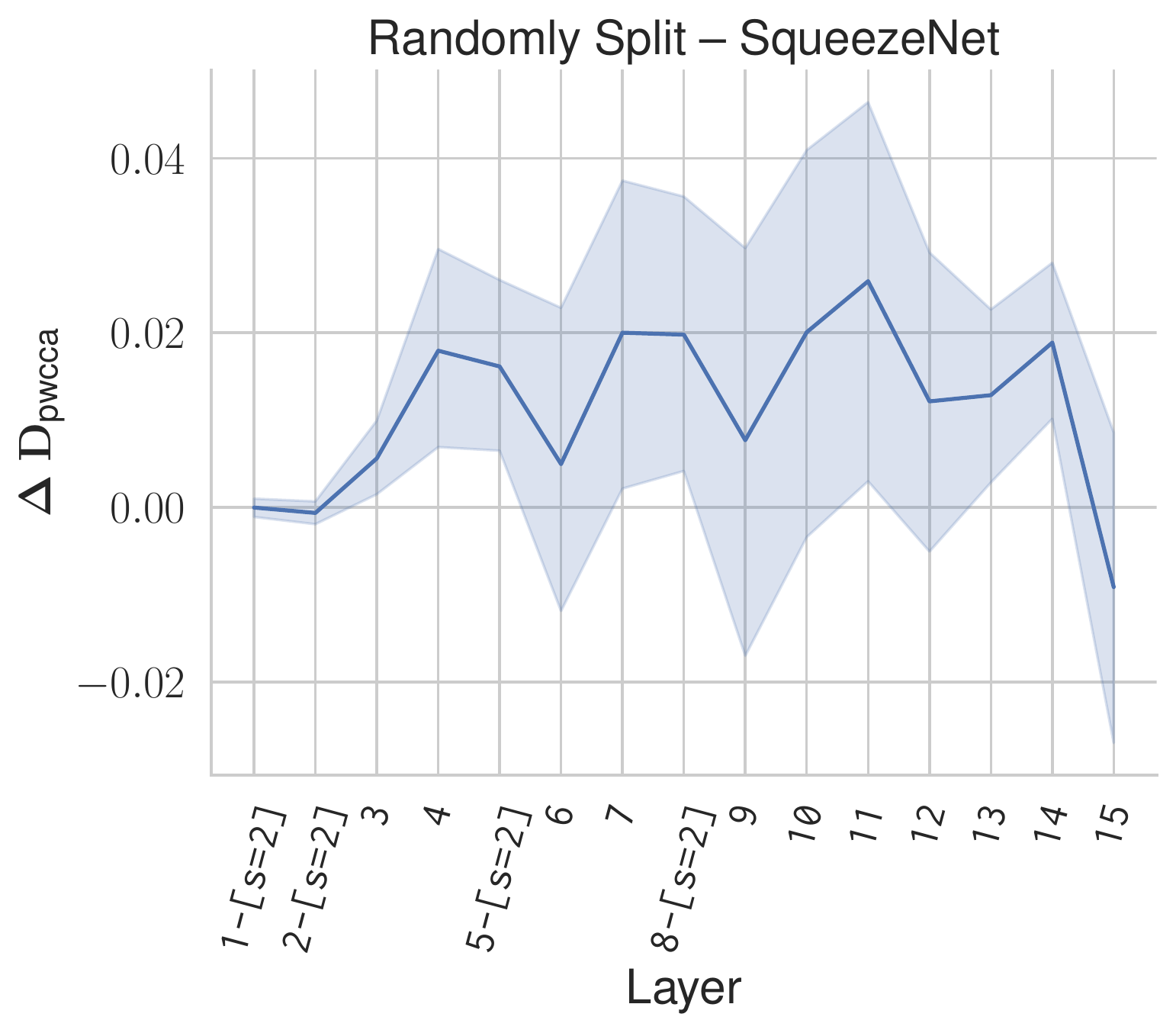}
    \\
    \includegraphics[width=0.23\textwidth]{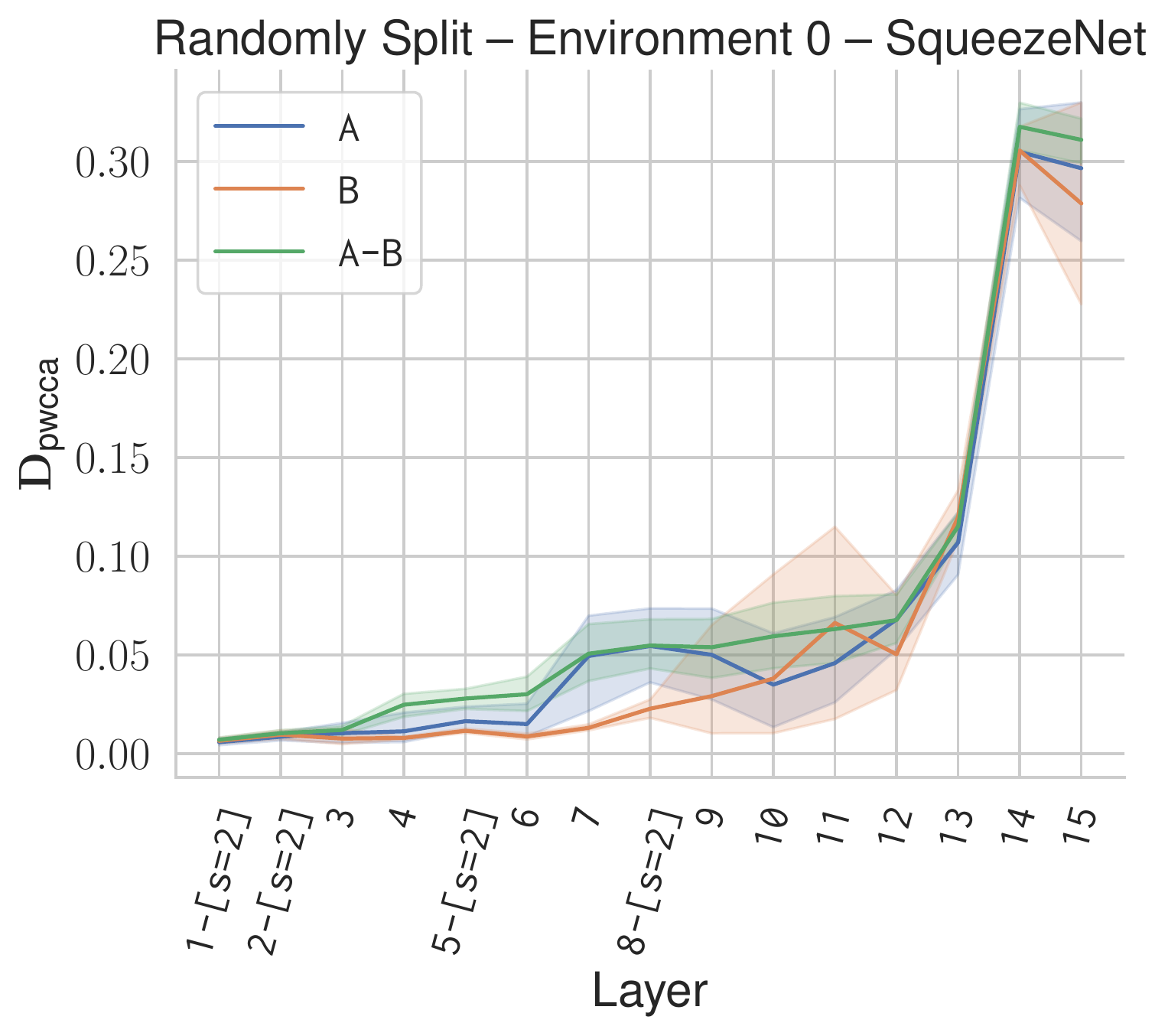}
    \includegraphics[width=0.23\textwidth]{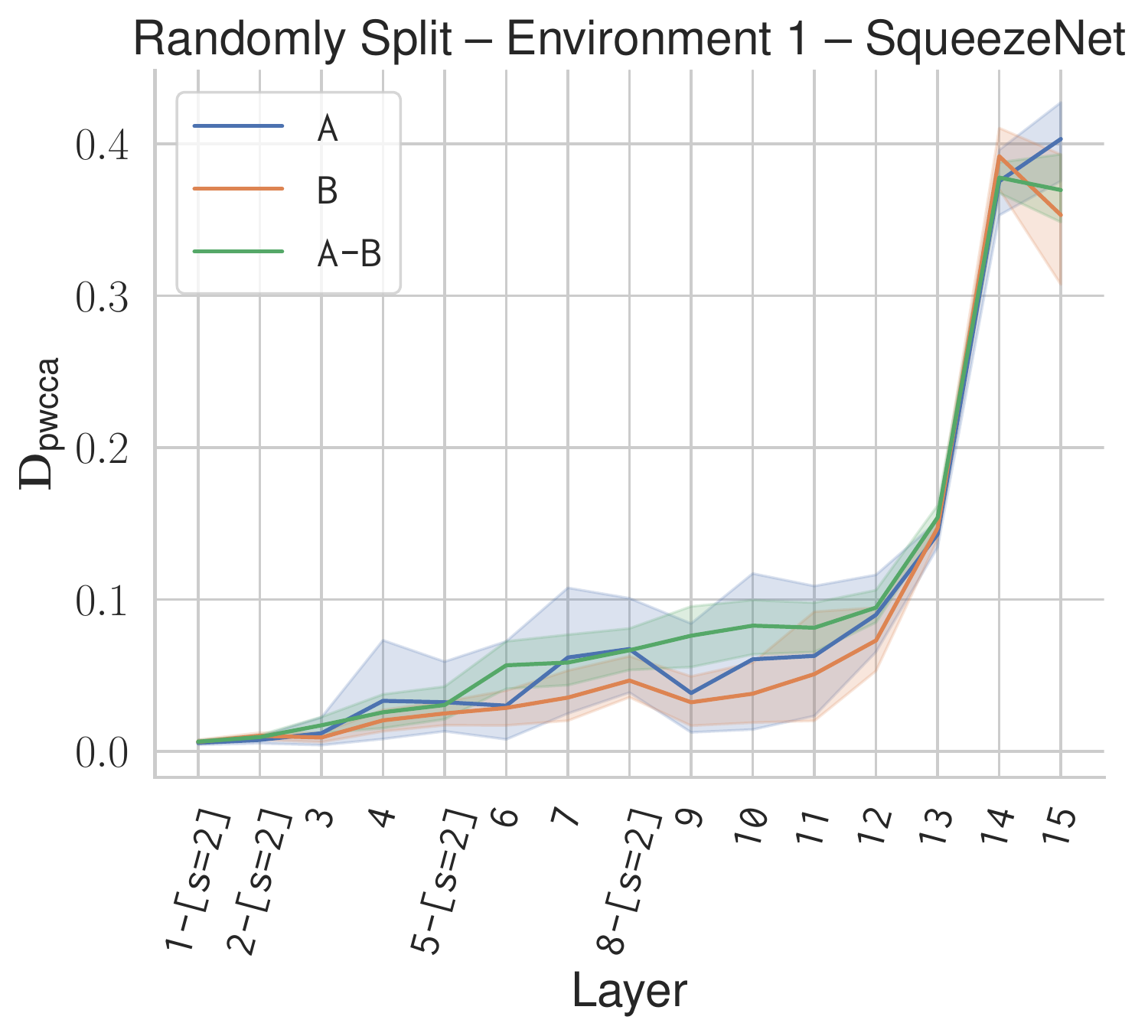}
    \includegraphics[width=0.23\textwidth]{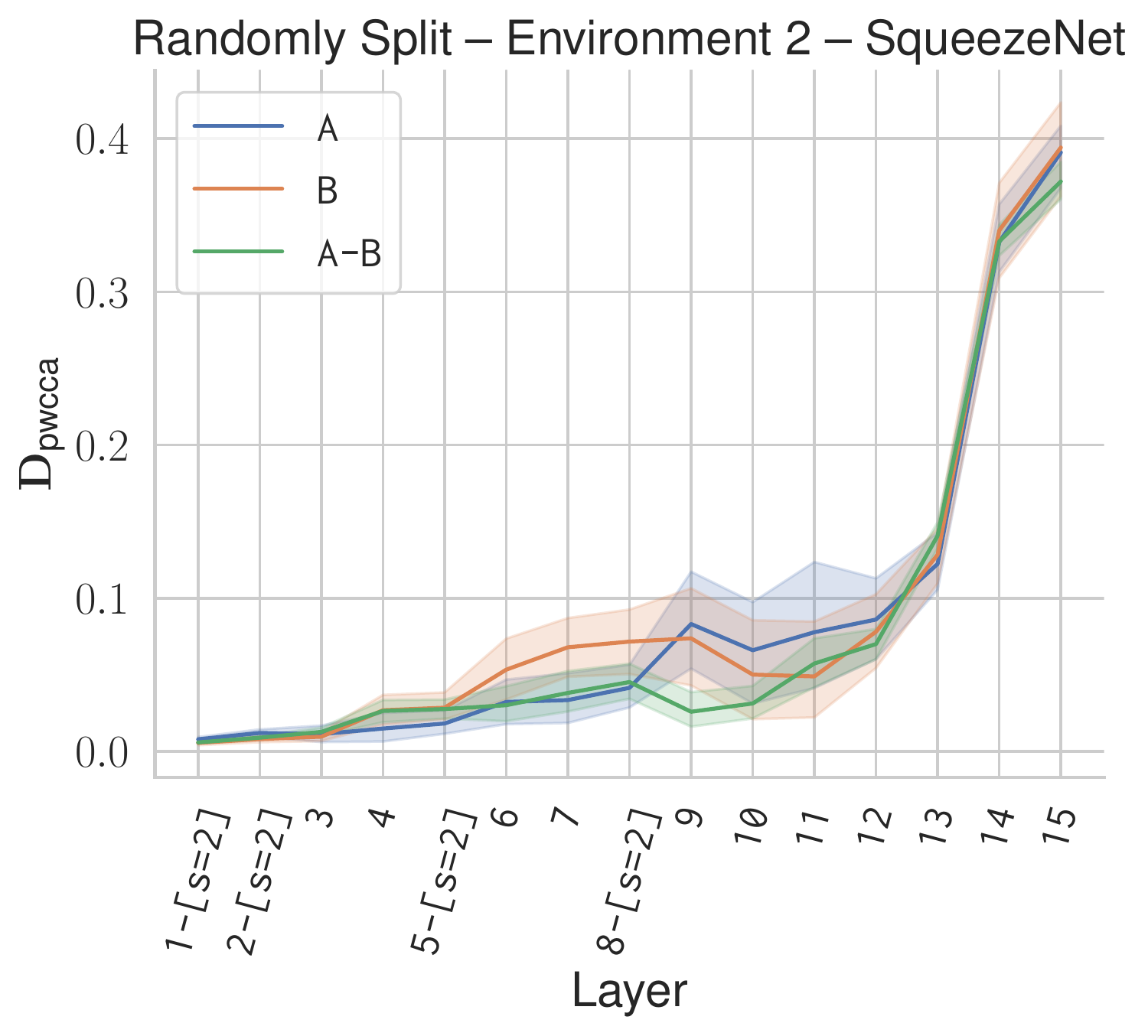}
    \includegraphics[width=0.23\textwidth]{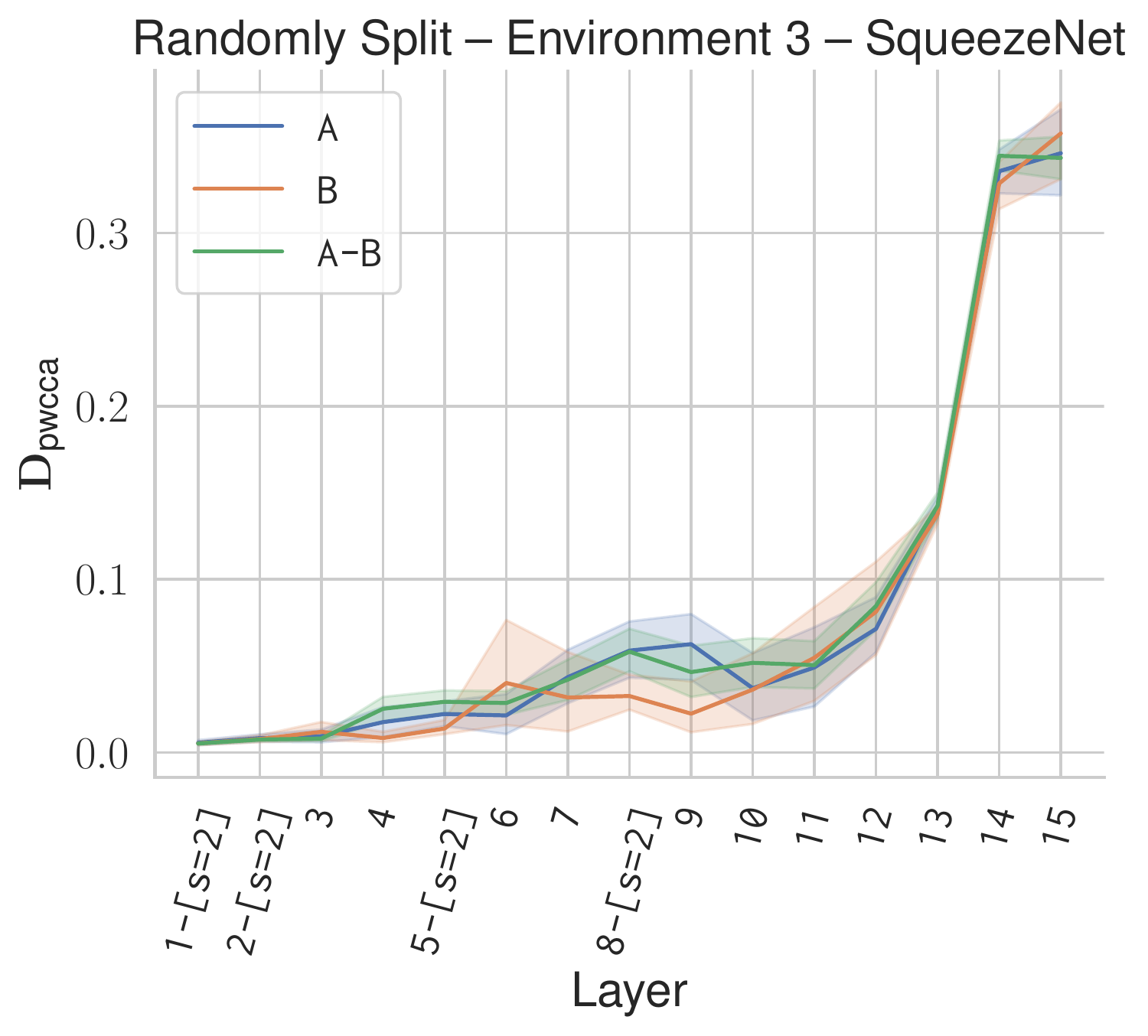}
    \caption{SqueezeNet results for randomly split sets of target objects on 4 environments from the replica dataset.  First plot shows the average $\mathbf{\Delta D_{\text{pwcca}}} = \mathbf{D_{\text{pwcca}}}(\text{A-B}) - (\mathbf{D_{\text{pwcca}}}(\text{A}) - \mathbf{D_{\text{pwcca}}}(\text{B}))/2.0$ across all environments.}
    \label{fig:sqz_random_replica}
\end{figure}

\begin{figure}
    \centering
    \includegraphics[width=0.5\textwidth]{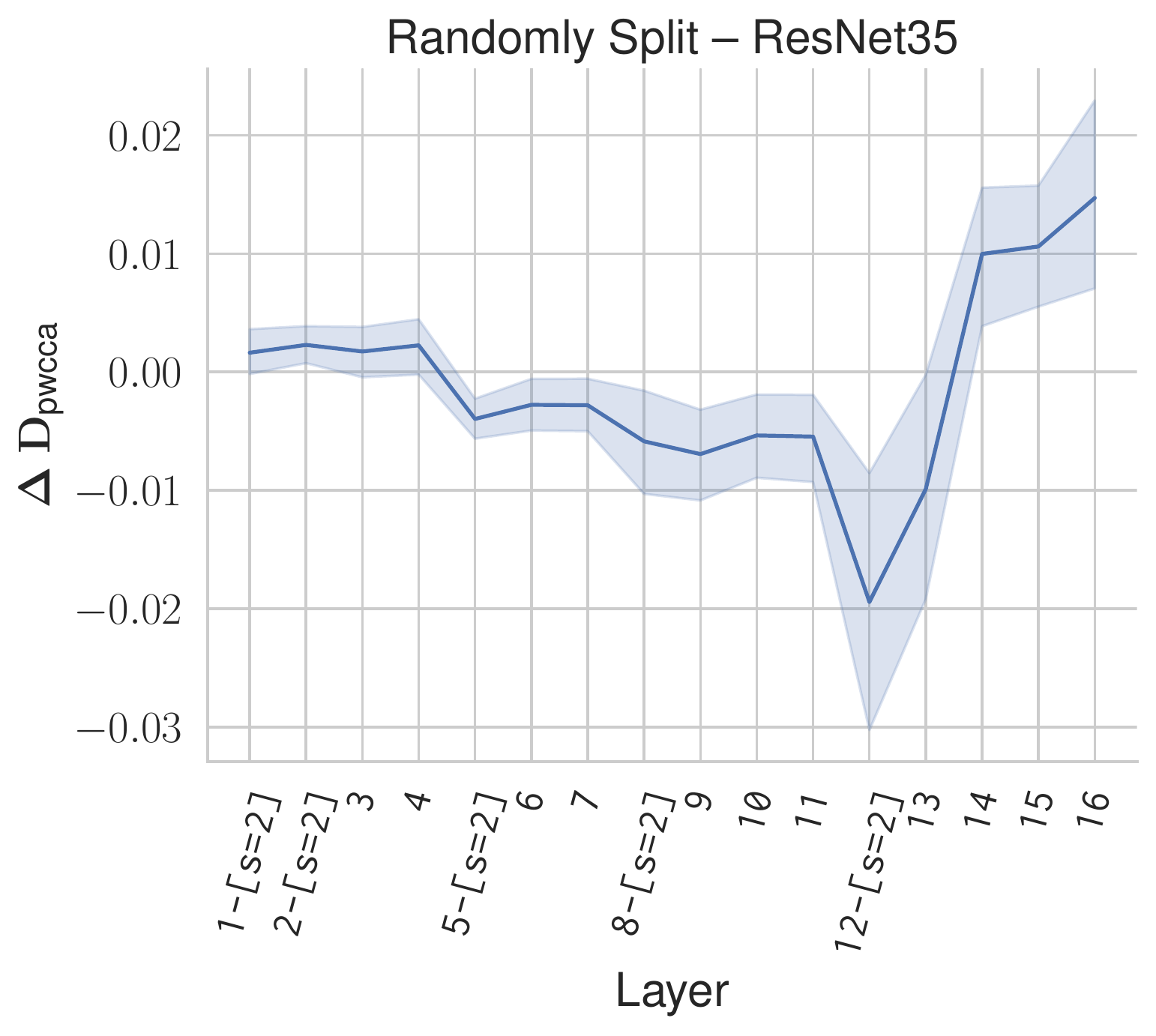}
    \\
    \includegraphics[width=0.23\textwidth]{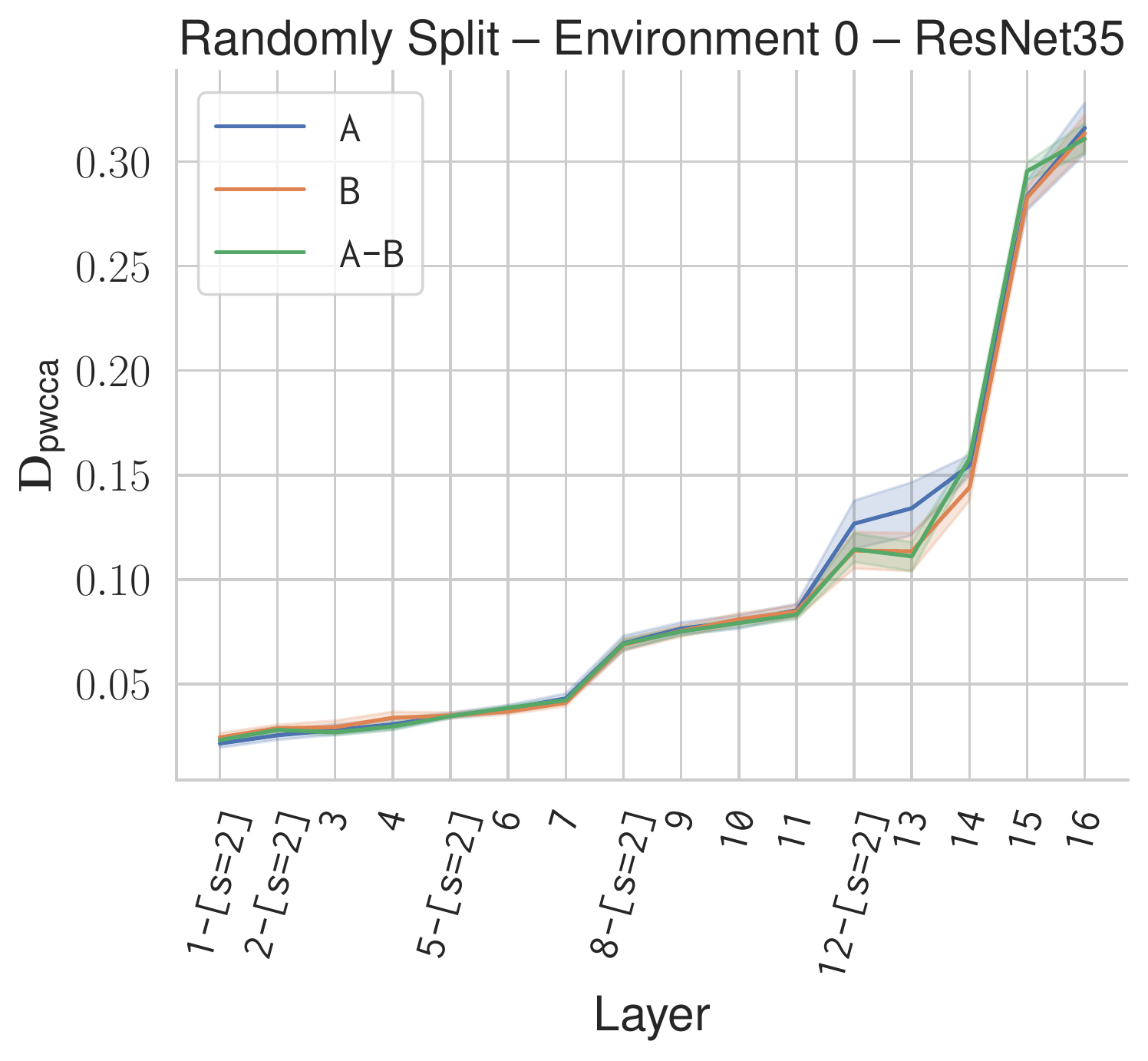}
    \includegraphics[width=0.23\textwidth]{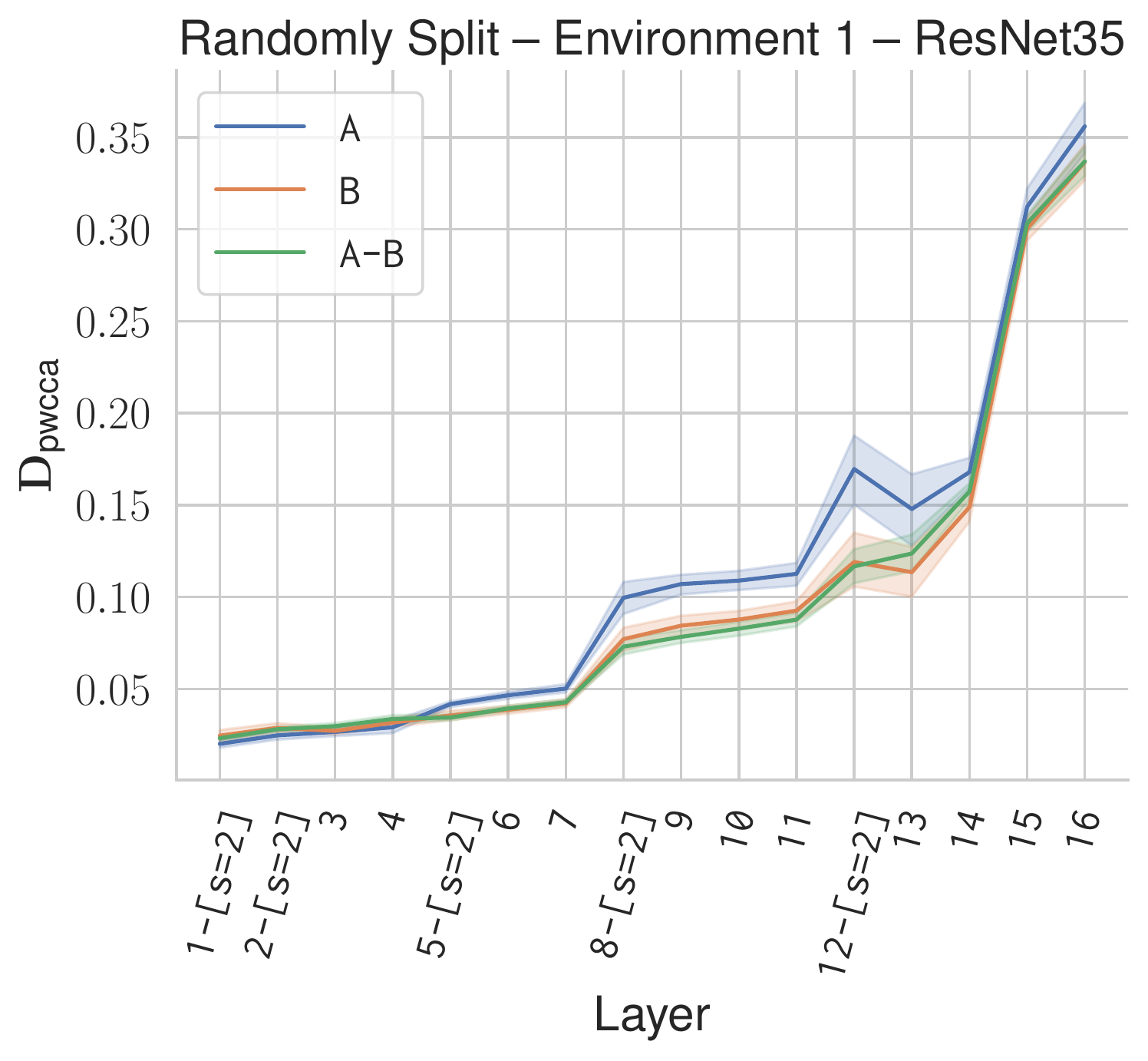}
    \includegraphics[width=0.23\textwidth]{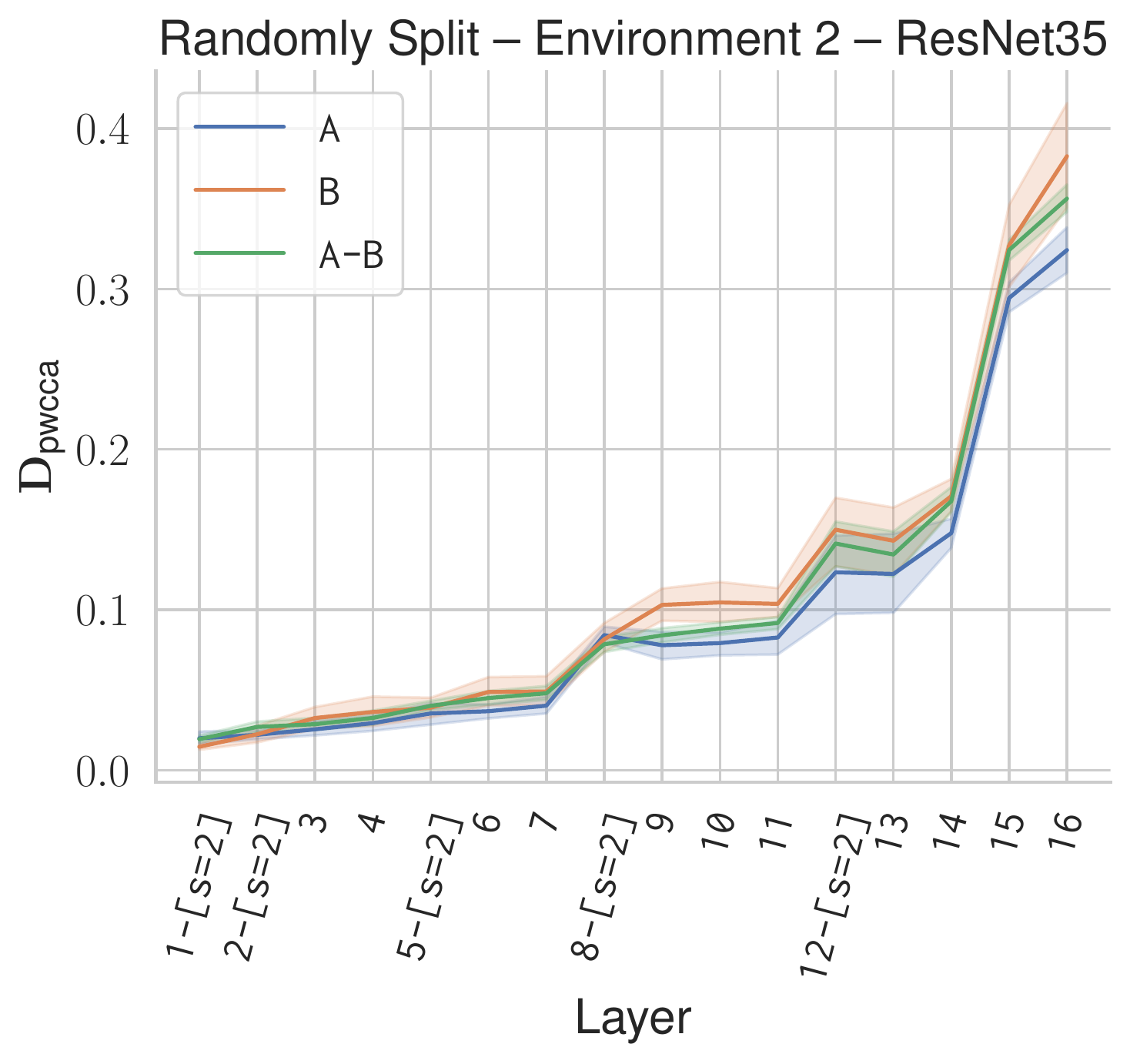}
    \includegraphics[width=0.23\textwidth]{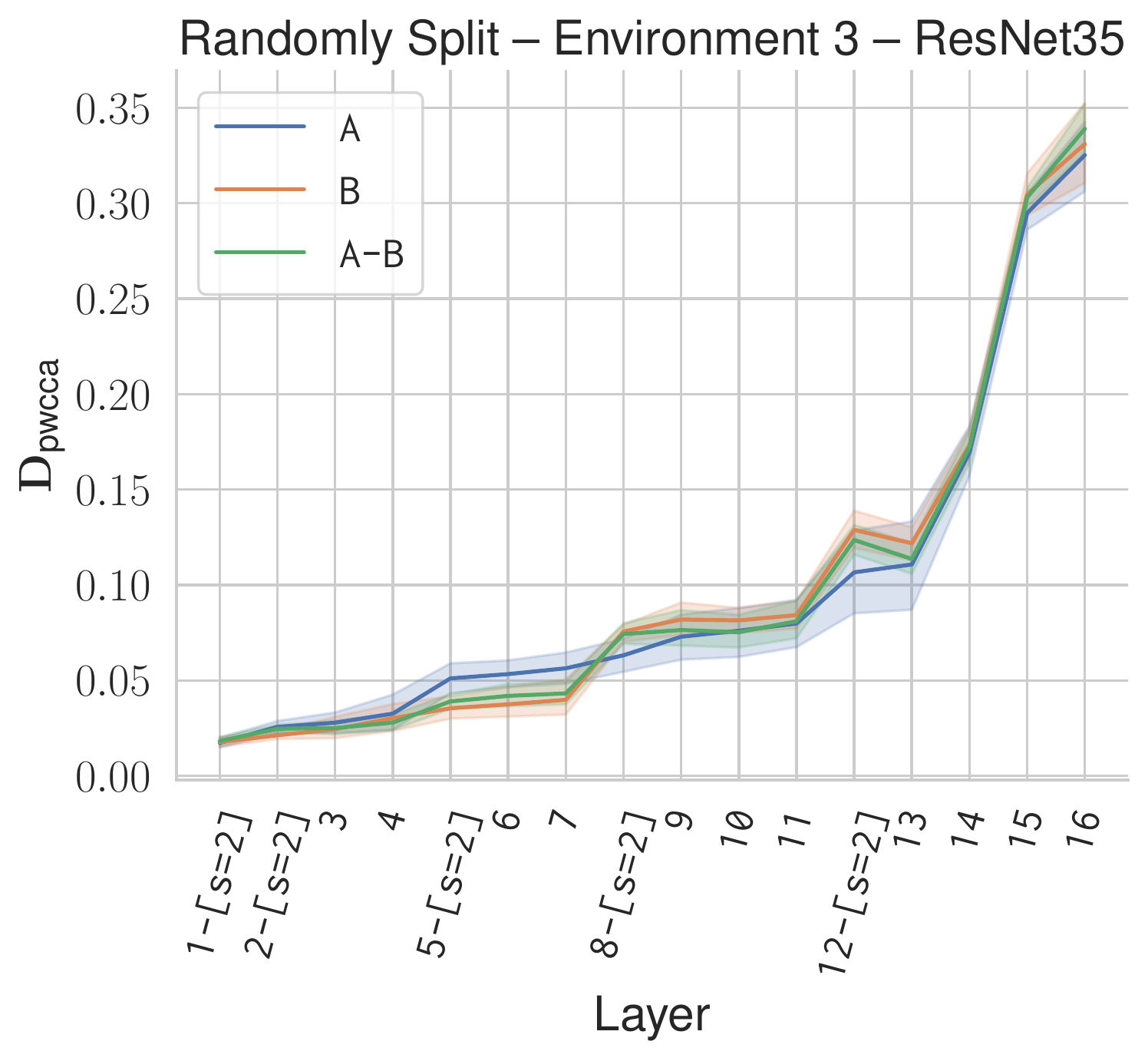}
    \caption{ResNet35 results for randomly split sets of target objects on 4 environments from the replica dataset.   First plot shows the average $\mathbf{\Delta D_{\text{pwcca}}} = \mathbf{D_{\text{pwcca}}}(\text{A-B}) - (\mathbf{D_{\text{pwcca}}}(\text{A}) - \mathbf{D_{\text{pwcca}}}(\text{B}))/2.0$ across all environments.}
    \label{fig:rs35_random_replica}
\end{figure}